\documentclass[10pt,journal,compsoc]{IEEEtran}
\ifCLASSOPTIONcompsoc
  \usepackage[nocompress]{cite}
\else
  \usepackage{cite}
\fi
\ifCLASSINFOpdf
\else
\fi
\usepackage{epsfig}
\usepackage{graphicx}
\usepackage{amsmath}
\usepackage{amssymb}

\usepackage{epsfig}
\usepackage{graphicx}
\usepackage{algorithm}
\usepackage{algorithmic}

\def\E{{\rm E}}

\def\KL{{\rm KL}}

\def\q{q_{\rm data}}

\def\D{{\cal D}}

\begin{document}
%
\title{Divergence Triangle for Joint Training of Generator Model, Energy-based Model, and Inference Model}
%
%
%
%
\newcommand*\samethanks[1][\value{footnote}]{\footnotemark[#1]}
\author{Tian Han*\thanks{* Equal contributions.},
        Erik Nijkamp*, 
        Xiaolin Fang,
        Mitch Hill,
        Song-Chun Zhu,
        Ying Nian Wu\\
        Department of Statistics, UCLA}
\IEEEtitleabstractindextext{%
\begin{abstract}
This paper proposes the divergence triangle as a framework for joint training of generator model, energy-based model and inference model. The divergence triangle is a compact and symmetric (anti-symmetric) objective function that seamlessly integrates variational learning, adversarial learning, wake-sleep algorithm, and contrastive divergence in a unified probabilistic formulation. This unification makes the processes of sampling, inference, energy evaluation readily available without the need for costly Markov chain Monte Carlo methods. Our experiments demonstrate that the divergence triangle is capable of learning (1) an energy-based model with well-formed energy landscape, (2) direct sampling in the form of a generator network, and (3) feed-forward inference that faithfully reconstructs observed as well as synthesized data. The divergence triangle is a robust training method that can learn from incomplete data.
\end{abstract}

}

\maketitle

\IEEEdisplaynontitleabstractindextext

%
\IEEEpeerreviewmaketitle

\IEEEraisesectionheading{\section{Introduction}\label{sec:introduction}}

\subsection{Integrating Three Models}

Deep probabilistic generative models are a powerful framework for representing complex data distributions. They have been widely used in unsupervised learning problems to learn from unlabeled data. The goal of generative learning is to build rich and flexible models to fit complex, multi-modal data distributions as well as to be able to generate samples with high realism. The family of generative models may be roughly divided into two classes: The first class is the \textit{energy-based model} (a.k.a undirected graphical model) and the second class is the latent variable model (a.k.a directed graphical model) which usually includes \textit{generator model} for the generation and \textit{inference model} for inference or reconstruction. 

These models have their advantages and limitations. An energy-based model defines an explicit likelihood of the observed data up to a normalizing constant. However, sampling from such a model usually requires expensive Markov chain Monte Carlo (MCMC). A generator model defines direct sampling of the data. However, it does not have an explicit likelihood. The inference of the latent variables also requires MCMC sampling from the posterior distribution. The inference model defines an explicit approximation to the posterior distribution of the latent variables. 

Combining the energy-based model, the generator model, and the inference model to get the best of each model is an attractive goal. On the other hand, challenges may accumulate when the models are trained together since different models need to effectively compete or cooperate together to achieve their highest performances. In this work, we propose the divergence triangle for joint training of energy-based model, generator model and inference model. The learning of three models can then be seamlessly integrated in a principled probabilistic framework. The energy-based model is learned based on the samples supplied by the generator model. With the help of the inference model, the generator model is trained by both the observed data and the energy-based model. The inference model is learned from both the real data fitted by the generator model as well as the synthesized data generated by the generator model.

Our experiments demonstrate that the divergence triangle is capable of learning an energy-based model with a well-behaved energy landscape, a generator model with highly realistic samples, and an inference model with faithful reconstruction ability.

\subsection{Prior Art}

The divergence triangle jointly learns an energy-based model, a generator model, and an inference model. The following are previous methods for learning such models. 

The maximum likelihood learning of the energy-based model requires expectation with respect to the current model, while the maximum likelihood learning of the generator model requires expectation with respect to the posterior distribution of the latent variables. Both expectations can be approximated by MCMC, such as Gibbs sampling~\cite{gibbs}, Langevin dynamics, or Hamiltonian Monte Carlo (HMC)~\cite{neal2011mcmc}. \cite{LuZhuWu2016, xieLuICML} used Langevin dynamics for learning the energy-based models, and \cite{HanLu2016}  used Langevin dynamics for learning the generator model. In both cases, MCMC sampling introduces an inner loop in the training procedure, posing a computational expense.

An early version of the energy-based model is the FRAME (Filters, Random field, And Maximum Entropy) model \cite{zhu1997minimax, wu2000equivalence}.  \cite{zhu1997GRADE} used gradient-based method such as Langevin dynamics to sample from the model. \cite{zhu2003statistical} called the energy-based models as descriptive models. \cite{LuZhuWu2016, xieLuICML} generalized the model to deep variants.

For learning the energy-based model \cite{lecun2006tutorial}, to reduce the computational cost of MCMC sampling, contrastive divergence (CD)~\cite{hinton} initializes a finite step MCMC from the observed data. The resulting learning algorithm follows the gradient of the difference between two Kullback-Leibler divergences, thus the name contrastive divergence. In this paper, we shall use the term ``contrastive divergence'' in a more general sense than \cite{hinton}. Persistent contrastive divergence~\cite{pcd} initializes MCMC sampling from the samples of the previous learning iteration.

Generalizing \cite{tu2007learning}, \cite{TuNIPS} developed an introspective learning method where the energy function is discriminatively learned, and the energy-based model is both a generative model and a discriminative model. 

For learning the generator model, the variational auto-encoder (VAE)~\cite{kingma2013auto, RezendeICML2014, MnihGregor2014} approximates the posterior distribution of the latent variables by an explicit inference model. In VAE, the inference model is learned jointly with the generator model from the observed data. A precursor of VAE is the wake-sleep algorithm~\cite{hinton1995wake}, where the inference model is learned from the dream data generated by the generator model in the sleep phase. 

The generator model can also be learned jointly with a discriminator model,  as in the generative adversarial networks (GAN)~\cite{goodfellow2014generative}, as well as deep convolutional GAN (DCGAN)~\cite{radford2015unsupervised}, energy-based GAN (EB-GAN)~\cite{zhao2016energy}, Wasserstein GAN (WGAN)~\cite{arjovsky2017wasserstein}. GAN does not involve an inference model. 

The generator model can also be learned jointly with an energy-based model \cite{Bengio2016, dai2017calibrating}. We can interpret the learning scheme as an adversarial version of contrastive divergence. While in GAN, the discriminator model eventually becomes a confused one, in the joint learning of the generator model and the energy-based model, the learned energy-based model becomes a well-defined probability distribution on the observed data. The joint learning bares some similarity to WGAN, but unlike WGAN, the joint learning involves two complementary probability distributions. 

To bridge the gap between the generator model and the energy-based model, the cooperative learning method of \cite{coopnets_pami} introduces finite-step MCMC sampling of the energy-based model with the MCMC initialized from the samples generated by the generator model. Such finite-step MCMC produces synthesized examples closer to the energy-based model, and the generator model can learn from how the finite-step MCMC revises its initial samples. 

Adversarially learned inference (ALI)~\cite{dumoulin2016adversarially,donahue2016adversarial} combines the learning of the generator model and inference model in an adversarial framework. ALI can be improved by adding conditional entropy regularization, resulting in the ALICE~\cite{li2017alice} model. The recently proposed method~\cite{chen2018symmetric} shares the same spirit. They lack an energy-based model on observed data.

\subsection{Our Contributions}

Our proposed formulation, which we call the \textit{divergence triangle}, re-interprets and integrates the following elements in unsupervised generative learning: (1) maximum likelihood learning, (2) variational learning, (3) adversarial learning, (4) contrastive divergence, (5) wake-sleep algorithm. The learning is seamlessly integrated into a probabilistic framework based on KL divergence.

We conduct extensive experiments to analyze the learned models. Energy landscape mapping is used to verify that our learned energy-based model is well-behaved. Further, we evaluate the learning of a generator model via synthesis by generating samples with competitive fidelity, and evaluate the accuracy of the inference model both qualitatively and quantitatively via reconstruction. Our proposed model can also benefit in learning directly from incomplete images with various blocking patterns.

\section{Learning Deep Probabilistic Models}

In this section, we shall review the two probabilistic models, namely the generator model and the energy-based model, both of which are parametrized by convolutional neural networks \cite{lecun1998gradient, krizhevsky2012imagenet}. Then, we shall present the maximum likelihood learning algorithms for training these two models, respectively. Our presentation of the two maximum likelihood learning algorithms is unconventional. We seek to derive both algorithms based on the Kullback-Leibler divergence using the same scheme. This will set the stage for the divergence triangle.

\subsection{Generator Model and Energy-based Model}

The generator model \cite{goodfellow2014generative, radford2015unsupervised, kingma2013auto, RezendeICML2014, MnihGregor2014} is a generalization of the factor analysis model \cite{rubin1982algorithms},
\begin{eqnarray} 
z \sim {\rm N}(0, I_d), \; x = g_\theta(z) + \epsilon,
\end{eqnarray} 
where $g_\theta$ is a top-down mapping parametrized by a deep network with parameters $\theta$.  It maps the $d$-dimensional latent vector $z$ to the $D$-dimensional signal $x$. $\epsilon \sim {\rm N}(0, \sigma^2 I_D)$ and is independent of $z$. In general, the model is defined by the prior distribution $p(z)$ and the conditional distribution $p_\theta(x|z)$. The complete-data model $p_\theta(z, x) = p(z) p_\theta(x|z)$. The observed-data model is $p_\theta(x) = \int p_\theta(z, x) dz$. The posterior distribution is $p_\theta(z|x) = p_\theta(z, x)/p_\theta(x)$. See the diagram (a) below. 
\begin{eqnarray*}
\begin{array}[c]{ccc}
	\mbox{{Top-down} mapping}  && \mbox{{Bottom-up} mapping}\\
	\mbox{{hidden vector} $z$} && \mbox{{energy} $-f_\alpha(x)$}\\
	\Downarrow&&\Uparrow\\
	\mbox{signal $x \approx g_\theta(z)$} && \mbox{signal $x$}\\
	\mbox{(a) Generator model} && \mbox{(b) Energy-based model}
\end{array}  \label{eq:diagram0}
\end{eqnarray*}

A complementary model is the energy-based model \cite{Ng2011, Dai2015ICLR, LuZhuWu2016, xieLuICML}, where $-f_\alpha(x)$ defines the energy of $x$, and a low energy $x$ is assigned a high probability. Specifically, we have the following probability model 
\begin{eqnarray}
\pi_\alpha(x) = \frac{1}{Z(\alpha)} \exp\left[ f_\alpha(x) \right], 
\end{eqnarray} 
where  $f_\alpha(x)$ is parametrized by a bottom-up deep network with parameters $\alpha$, and $Z(\alpha)$ is the normalizing constant. If $f_\alpha(x)$ is linear in $\alpha$, the model becomes the familiar exponential family model in statistics or the Gibbs distribution in statistical physics. We may consider $\pi_\alpha$ an evaluator, where $f_\alpha$ assigns the value to $x$, and $\pi_\alpha$ evaluates $x$ by a normalized probability distribution.  See the diagram (b) above. 

The energy-based model $\pi_\alpha$ defines explicit log-likelihood via $f_\alpha(x)$, even though $Z(\alpha)$ is intractable. However, it is difficult to sample from $\pi_\alpha$. The generator model $p_\theta$ can generate $x$ directly by first generating $z \sim p(z)$, and then transforming $z$ to $x$ by $g_\theta(z)$. But it does not define an explicit log-likelihood of $x$. 

In the context of inverse reinforcement learning \cite{ziebart2008maximum, abbeel2004apprenticeship} or inverse optimal control, $x$ is action and $-f_\alpha(x)$ defines the cost function or $f_\alpha(x)$ defines the value function or the objective function.

\subsection{Maximum Likelihood Learning}

Let $q_{\rm data}(x)$ be the true distribution that generates the training data. Both the generator $p_\theta$ and the energy-based model $\pi_\alpha$ can be learned by maximum likelihood. For large sample, the maximum likelihood amounts to minimizing the Kullback-Leibler divergence ${\rm KL}(q_{\rm data}\|p_\theta)$ over $\theta$, and minimizing ${\rm KL}(q_{\rm data}\|\pi_\alpha)$ over $\alpha$, respectively. The expectation $\E_{q_{\rm data}}$ can be approximated by sample average. 

\subsubsection{EM-type Learning of Generator Model}

To learn the generator model $p_\theta$, we seek to minimize ${\rm KL}(q_{\rm data}(x)\|p_\theta(x))$ over $\theta$. Suppose in an iterative algorithm, the current $\theta$ is $\theta_t$. We can fix $\theta_t$ at any place we want, and vary $\theta$ around $\theta_t$. 

We can write 
\begin{eqnarray}
&&{\rm KL}(q_{\rm data}(x) p_{\theta_t}(z|x) \|p_\theta(z, x))=\nonumber\\
&&{\rm KL}(q_{\rm data}(x) \|p_\theta(x)) + {\rm KL}(p_{\theta_t}(z|x)\|p_\theta(z|x)). \label{eq:VAE0}
\end{eqnarray}
In the EM algorithm \cite{dempster1977maximum}, the left hand side is the surrogate objective function. This surrogate function is more tractable than the true objective function $ {\rm KL}(q_{\rm data}(x) \|p_\theta(x))$ because $q_{\rm data}(x) p_{\theta_t}(z|x)$ is a distribution of the complete data, and $p_\theta(z, x)$ is the complete-data model. 

\begin{figure}[ht]
	\centering	
	\includegraphics[width=.43\linewidth]{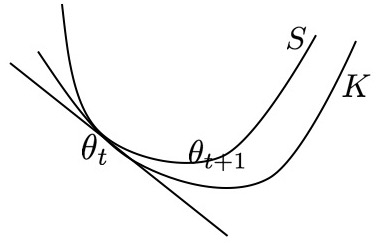} 
	\caption{\small The surrogate $S$ majorizes (upper bounds) $K$, and they touch each other at $\theta_t$ with the same tangent. }	
	\label{fig:k1}
\end{figure}

We can write (\ref{eq:VAE0}) as 
\begin{eqnarray} 
S(\theta) = K(\theta) + \tilde{K}(\theta). \label{eq:v0}
\end{eqnarray}
The geometric picture is that the surrogate objective function $S(\theta)$ is above the true objective function $K(\theta)$, i.e., $S$ majorizes (upper bounds) $K$, and they touch each other at $\theta_t$, so that $S(\theta_t) = K(\theta_t)$ and $S'(\theta_t) = K'(\theta_t)$. The reason is that $\tilde{K}(\theta_t) = 0$ and $\tilde{K}'(\theta_t) = 0$. See Figure \ref{fig:k1}.

$q_{\rm data}(x) p_{\theta_t}(z|x)$ gives us the complete data. Each step of EM fits the complete-data model $p_\theta(z, x)$ by minimizing the surrogate $S(\theta)$, 
\begin{eqnarray} 
\theta_{t+1} = \arg\min_\theta {\rm KL}(q_{\rm data} (x) p_{\theta_t}(z|x) \| p_\theta(z, x)),
\end{eqnarray} 
which amounts to maximizing the complete-data log-likelihood. By minimizing $S$, we will reduce $S(\theta)$ relative to $\theta_t$, and we will reduce $K(\theta)$ even more, relative to $\theta_t$, because of the majorization picture. 

We can also use gradient descent to update $\theta$. Because $S'(\theta_t) = K'(\theta_t)$, and we can place $\theta_t$ anywhere, we have 
\begin{eqnarray} 
&&- \frac{\partial}{\partial \theta} {\rm KL}(q_{\rm data}(x)\|p_\theta(x)) \nonumber \\
&&= \E_{q_{\rm data}(x) p_\theta(z|x)} \left[\frac{\partial}{\partial \theta} \log p_\theta(z, x)\right]. 
\end{eqnarray} 
To implement the above updates, we need to compute the expectation with respect to the posterior distribution $p_\theta(z|x)$. It can be approximated by MCMC such as Langevin dynamics or HMC~\cite{neal2011mcmc}.  Both require gradient computations that can be efficiently accomplished by back-propagation. We have learned the generator using such learning method~\cite{HanLu2016}. 

\subsubsection{Self-critic Learning of Energy-based Model}

To learn the energy-based model $\pi_\alpha$, we seek to minimize ${\rm KL}(q_{\rm data}(x)\|\pi_\alpha(x))$ over $\alpha$. Suppose in an iterative algorithm, the current $\alpha$ is $\alpha_t$. We can fix $\alpha_t$ at any place we want, and vary $\alpha$ around $\alpha_t$.

Consider the following contrastive divergence
\begin{eqnarray}
{\rm KL}(q_{\rm data}(x)\|\pi_\alpha(x)) - {\rm KL}(\pi_{\alpha_t}(x)\|\pi_\alpha(x)). \label{eq:A0}
\end{eqnarray}
We can use the above as surrogate function, which is more tractable than the true objective function, since the $\log Z(\theta)$ term is canceled out. Specifically, we can write (\ref{eq:A0}) as 
\begin{eqnarray} 
S(\alpha) &=& K(\alpha) - \tilde{K}(\alpha)  \label{eq:a0} \\
&=& - (\E_{q_{\rm data}}[f_\alpha(x)] - \E_{\pi_{\alpha_t}}[f_\alpha(x)]) + {\rm const}. 
\end{eqnarray}

\begin{figure}[h]
	\centering	
	\includegraphics[width=.32\linewidth]{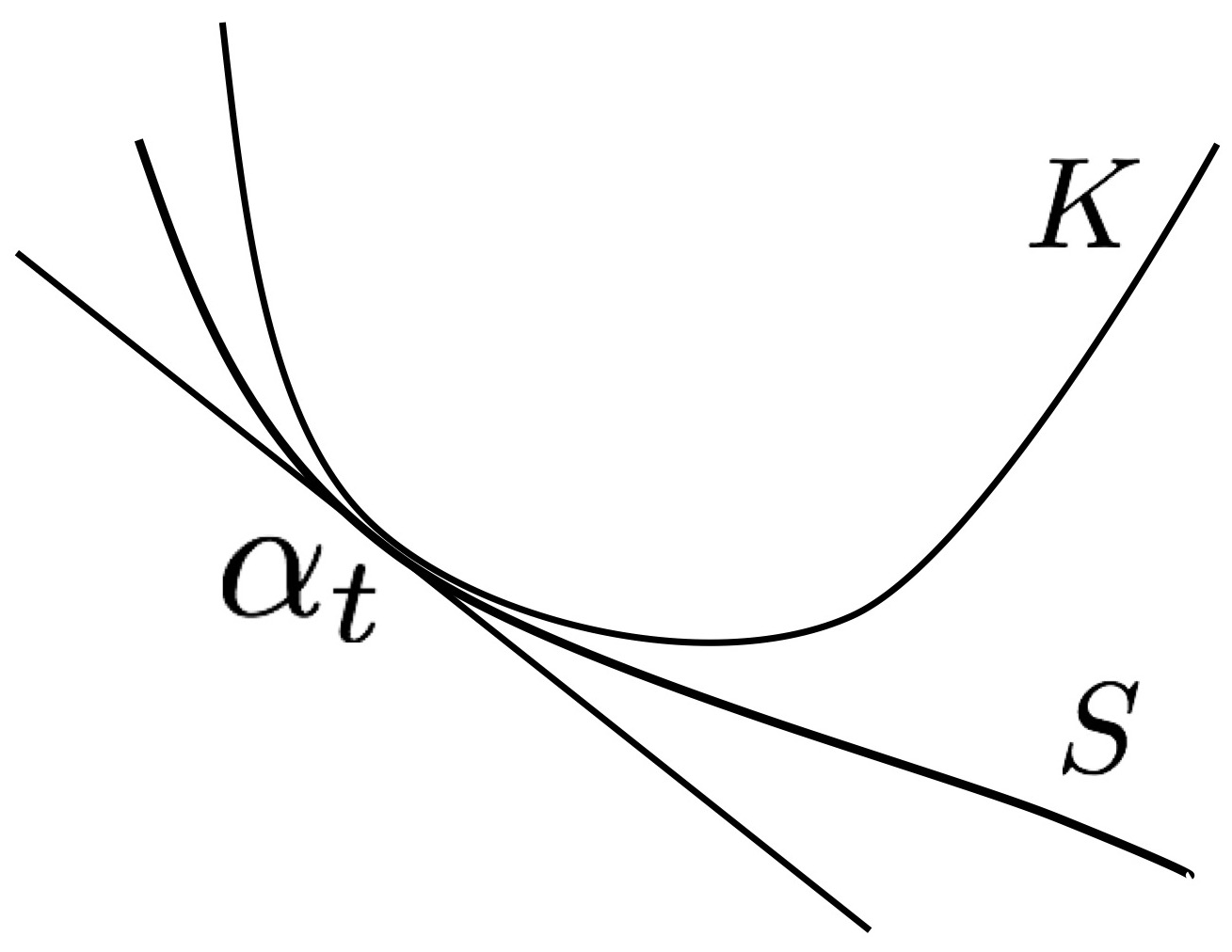} 
	\caption{\small The surrogate $S$ minorizes (lower bounds) $K$, and they touch each other at $\alpha_t$ with the same tangent. }	
	\label{fig:k2}
\end{figure}

The geometric picture is that the surrogate function $S(\alpha)$ is below the true objective function $K(\alpha)$, i.e., $S$ minorizes  (lower bounds) $K$, and they touch  each other at $\alpha_t$, so that $S(\alpha_t) = K(\alpha_t)$, and $S'(\alpha_t) = K'(\alpha_t)$. The reason is that ${\tilde{K}(\alpha_t) = 0}$ and ${\tilde{K}'(\alpha_t) = 0}$. See Figure \ref{fig:k2}. 

Because $S$ minorizes $K$, we do not have a EM-like update. However, we can still use gradient descent to update $\alpha$, where the derivative is 
\begin{eqnarray}
K'(\alpha_t) = S'(\alpha_t) = -(\E_{q_{\rm data}}[f'_{\alpha_t}(x)] -\E_{\pi_{\alpha_t}}[f'_{\alpha_t}(x)]), 
\end{eqnarray}
where 
\begin{eqnarray}
f'_{\alpha_t}(x) =\frac{ \partial} {\partial \alpha}f_\alpha(x) \Big|_{\alpha_t}.
\end{eqnarray}
Since we can place $\alpha_t$ anywhere, we have 
\begin{eqnarray} 
&&- \frac{\partial}{\partial \alpha} {\rm KL}(q_{\rm data}(x)\|\pi_\alpha(x)) \nonumber\\
&&= \E_{q_{\rm data}} \left[\frac{\partial}{\partial \alpha} f_\alpha(x)\right] -  \E_{\pi_\alpha} \left[\frac{\partial}{\partial \alpha} f_\alpha(x)\right].  \label{eq:e1}
\end{eqnarray} 
To implement the above update, we need to compute the expectation with respect to the current model $\pi_{\alpha_t}$. It can be approximated by MCMC such as Langevin dynamics or HMC that samples from $\pi_{\alpha_t}$. It can be efficiently implemented by gradient computation via back-propagation. We have trained the energy-based model using such learning method \cite{LuZhuWu2016, xieLuICML}. 

The above learning algorithm has an adversarial interpretation. Updating $\alpha_t$ to $\alpha_{t+1}$ by following the gradient of $S(\alpha) = {\rm KL}(q_{\rm data}(x)\|\pi_\alpha(x)) - {\rm KL}(\pi_{\alpha_t}(x)\|\pi_\alpha(x)) = -(\E_{q_{\rm data}}[f_\alpha(x)] - \E_{\pi_{\alpha_t}}[f_\alpha(x)]) + {\rm const}$, we seek to decrease the first KL-divergence, while we will increase the second KL-divergence, or we seek to shift the value function $f_\alpha(x)$ toward the observed data and away from the synthesized data generated from the current model. That is, the model $\pi_\alpha$ criticizes its current version $\pi_{\alpha_t}$, i.e., the model is its own adversary or its own critic. 

\subsubsection{Similarity and Difference}

In both models, at $\theta_t$ or $\alpha_t$, we have $S = K$, $S' = K'$, because $\tilde{K} = 0$ and $\tilde{K}' = 0$. 

The difference is that in the generator model, $S = K + \tilde{K}$, whereas in energy-based model, $S = K - \tilde{K}$. 

In the generator model, if we replace the intractable $p_{\theta_t}(z|x)$ by the inference model $q_\phi(z|x)$, we get VAE. 

In energy-based model, if we replace the intractable $\pi_{\alpha_t}(x)$ by the generator $p_\theta(x)$, we get adversarial contrastive divergence (ACD). The negative sign in front of $\tilde{K}$ is the root of the adversarial learning. 

\section{Divergence Triangle: Integrating Adversarial and Variational Learning}
In this section, we shall first present the divergence triangle, emphasizing its compact symmetric and anti-symmetric form. Then, we shall show that it is an re-interpretation and integration of existing methods, in particular, VAE~\cite{kingma2013auto, RezendeICML2014, MnihGregor2014} and ACD~\cite{Bengio2016, dai2017calibrating}.

\subsection{Loss Function}

Suppose we observe training examples $\{x_{(i)} \sim q_{\rm data}(x)\}_{i=1}^{n}$ where $q_{\rm data}(x)$ is the unknown data distribution. ${\pi_\alpha(x) \propto \exp[f_\alpha(x)]}$ with energy function $-f_\alpha$ denotes the energy-based model with parameters $\alpha$. The generator model $p(z)p_\theta(x|z)$ has parameters $\theta$ and latent vector $z$. It is trivial to sample the latent distribution $p(z)$ and the generative process is defined as $z\sim p(z)$,  $x \sim p_\theta(x|z)$.

The maximum likelihood learning algorithms for both the generator and energy-based model require MCMC sampling. We modify the maximum likelihood KL-divergences by proposing a divergence triangle criterion, so that the two models can be learned jointly without MCMC. In addition to the generator $p_\theta$ and energy-based model $\pi_\alpha$, we also include an inference model $q_\phi(z|x)$ in the learning scheme. Such an inference model is a key component in the variational auto-encoder \cite{kingma2013auto, RezendeICML2014, MnihGregor2014}. The inference model $q_\phi(z|x)$ with parameters $\phi$ maps from the data space to latent space. In the context of EM, $q_\phi(z|x)$ can be considered an imputor that imputes the missing data $z$ to get the complete data $(z, x)$. 

The three models above define joint distributions over $z$ and $x$ from different perspectives. The two marginals, i.e., empirical data distribution $\q(x)$ and latent prior distribution $p(z)$, are known to us. The goal is to harmonize the three joint distributions so that the competition and cooperation between different loss terms improves learning.

\begin{figure}
	\centering	
	\includegraphics[width=.48\linewidth]{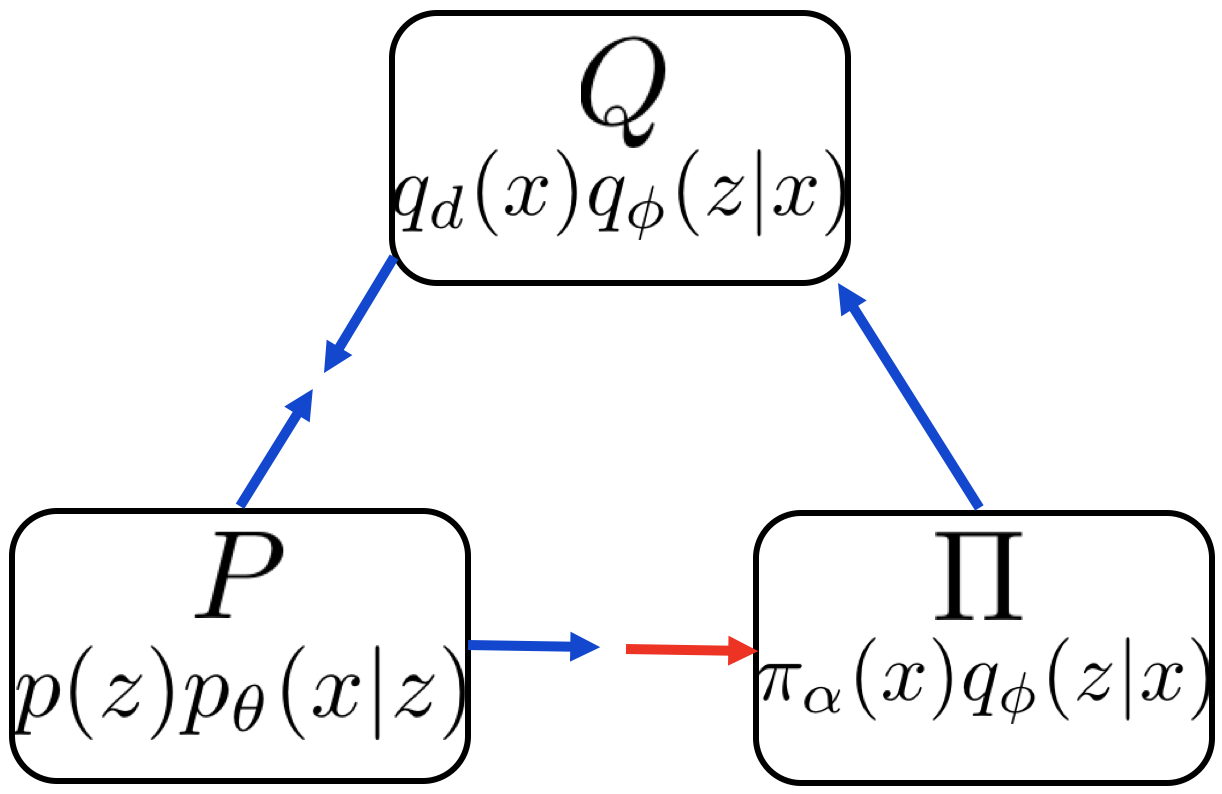} 
	\caption{\small Divergence triangle is based on the Kullback-Leibler divergences between three joint distributions of $(z, x)$. The blue arrow indicates the ``running toward'' behavior and the red arrow indicates the ``running away'' behavior.}	
	\label{fig:t1}
\end{figure}

The divergence triangle involves the following three joint distributions on $(z, x)$:
\begin{enumerate}
	\item $Q$-distribution: $Q(z, x) = q_{\rm data}(x) q_\phi(z|x)$.
	\item $P$-distribution: $P(z, x) = p(z) p_\theta(x|z)$.
	\item $\Pi$-distribution: $\Pi(z, x) = \pi_\alpha(x) q_\phi(z|x)$. 
\end{enumerate}

We propose to learn the three models $p_\theta$, $\pi_\alpha$, $q_\phi$ by the following divergence triangle loss functional $\D$
\begin{eqnarray} 
&& \max_\alpha \min_\theta \min_\phi \D(\alpha, \theta, \phi), \nonumber \\
&& \D = {\rm KL}(Q\|P) + {\rm KL}(P\|\Pi) - {\rm KL}(Q\|\Pi). \label{eq:triangle}
\end{eqnarray} 

See Figure \ref{fig:t1} for illustration. The divergence triangle is based on the three KL-divergences between the three joint distributions on $(z, x)$. It has a symmetric and anti-symmetric form, where the anti-symmetry is due to the negative sign in front of the last KL-divergence and the maximization over $\alpha$. The divergence triangle leads to the following dynamics between the three models: (1) $Q$ and $P$ seek to get close to each other. (2) $P$ seeks to get close to $\Pi$.  (3) $\pi$ seeks to get close to $q_{\rm data}$, but it seeks to get away from $P$, as indicated by the red arrow. Note that  $\KL(Q\|\Pi) = \KL(q_{\rm data}\|\pi_\alpha)$, because $q_\phi(z|x)$ is canceled out. The effect of (2) and (3) is that $\pi$ gets close to $q_{\rm data}$, while inducing $P$ to get close to $q_{\rm data}$ as well, or in other words, $P$ chases $\pi_\alpha$ toward $q_{\rm data}$. 

\subsection{Unpacking the Loss Function}

The divergence triangle integrates variational and adversarial learning methods, which are modifications of maximum likelihood. 

\subsubsection{Variational Learning}

 \begin{figure}[h]
	\centering	
	\includegraphics[height=.28\linewidth]{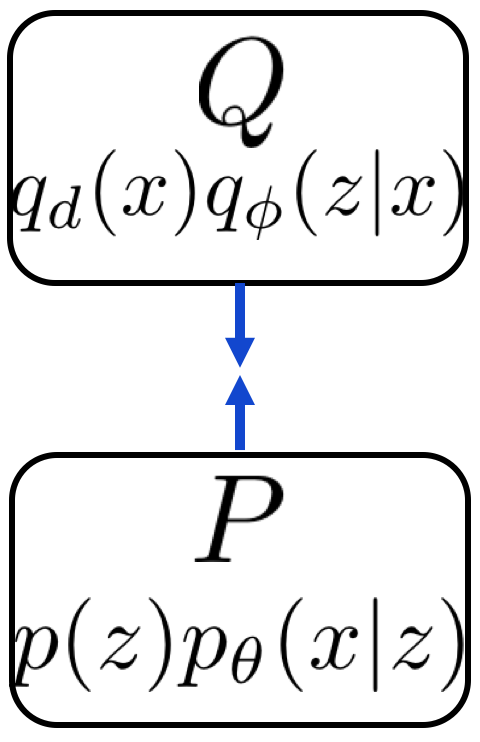} \includegraphics[height=.28\linewidth]{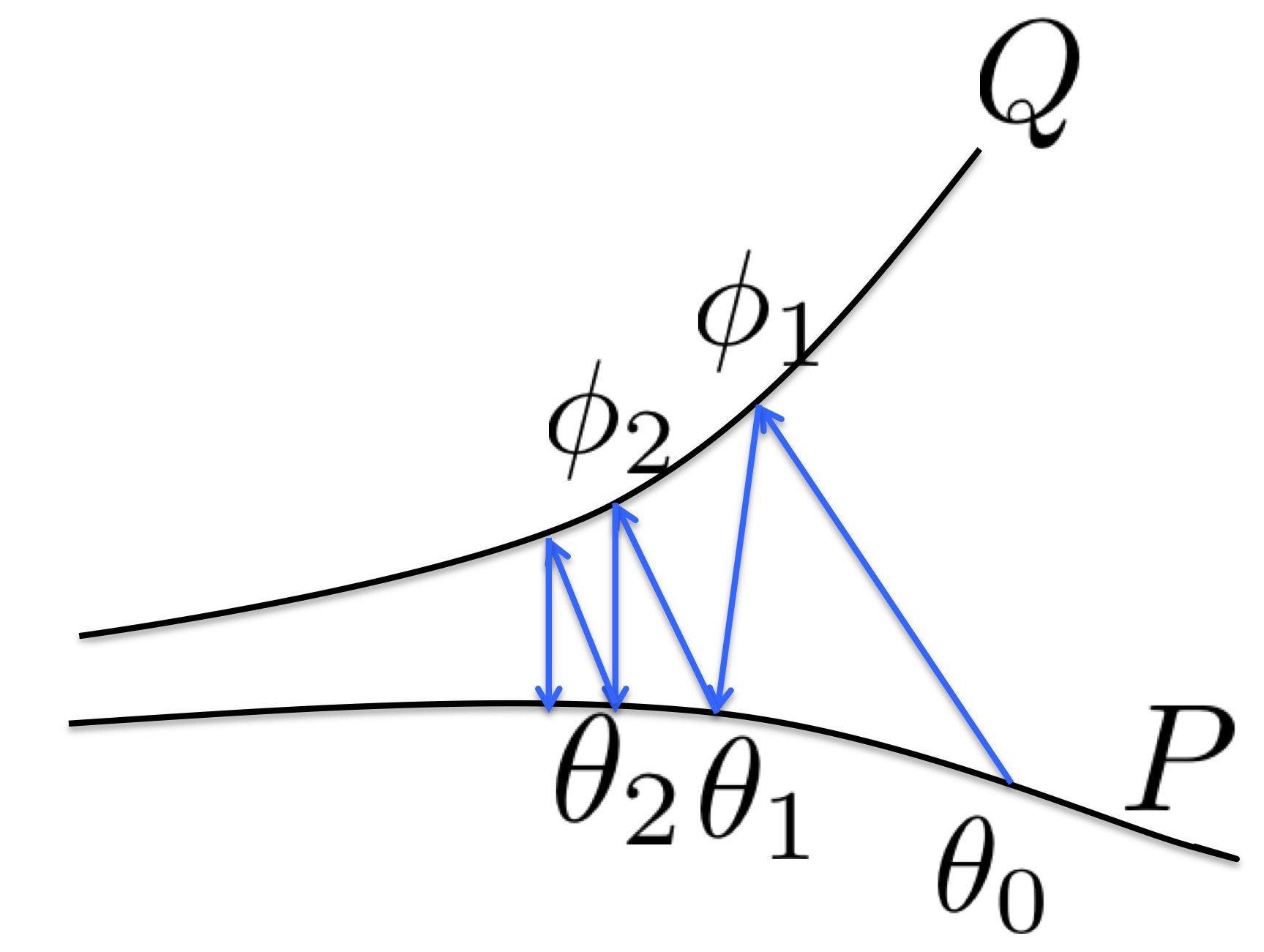} 
	\caption{\small Variational auto-encoder (VAE) as joint minimization by alternating projection. Left:  Interaction between the models. Right: Alternating projection. The two models run toward each other. }	
	\label{fig:t2}
\end{figure}   

First, $\min_\theta \min_\phi {\rm KL}(Q\|P)$ captures the variational auto-encoder (VAE). 
\begin{eqnarray}
{\rm KL}(Q\|P) &=& {\rm KL}(q_{\rm data}(x) \| p_\theta(x)) \nonumber\\
&+& {\rm KL}(q_\phi(z|x)\|p_\theta(z|x)), \label{eq:VAE}
\end{eqnarray}
Recall $S = K + \tilde{K}$ in (\ref{eq:v0}), if we replace the intractable $p_{\theta_t}(z|x)$ in (\ref{eq:v0}) by the explicit $q_\phi(z|x)$, we get (\ref{eq:VAE}), so that we avoid MCMC for sampling $p_{\theta_t}(z|x)$. 

We may interpret VAE as alternating projection between $Q$ and $P$. See Figure \ref{fig:t2} for illustration. If $q_\phi(z|x) = p_\theta(z|x)$, the algorithm reduces to the EM algorithm. The wake-sleep algorithm \cite{hinton1995wake} is similar to VAE, except that it updates $\phi$ by $\min_\phi {\rm KL}(P\|Q)$ instead of $\min_\phi \KL(Q\|P)$, so that the wake-sleep algorithm does not have a single objective function. 

The VAE $\min_\theta \min_\phi {\rm KL}(Q\|P)$ defines a cooperative game, with the dynamics that $q_\phi$ and $p_\theta$ run toward each other. 

\subsubsection{Adversarial Learning}

\begin{figure}[h]
	\centering	
	\includegraphics[height=.28\linewidth]{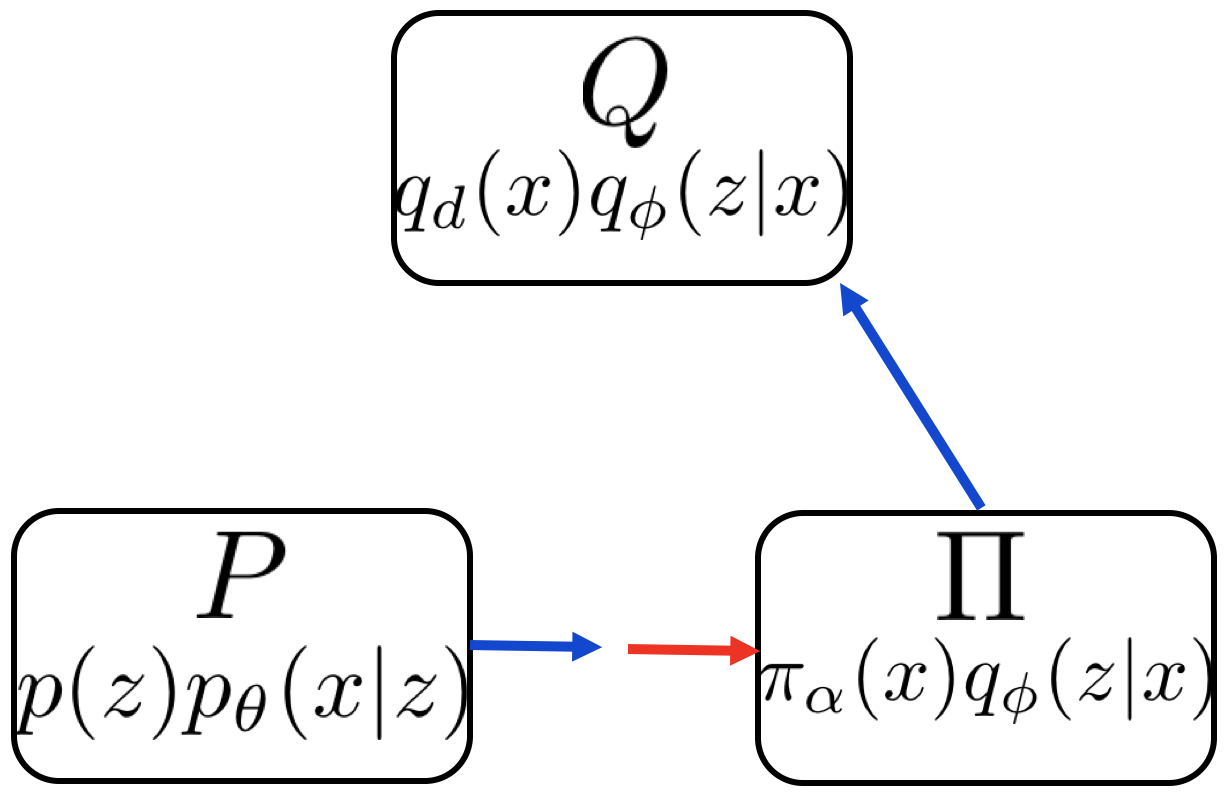} \includegraphics[height=.14\linewidth]{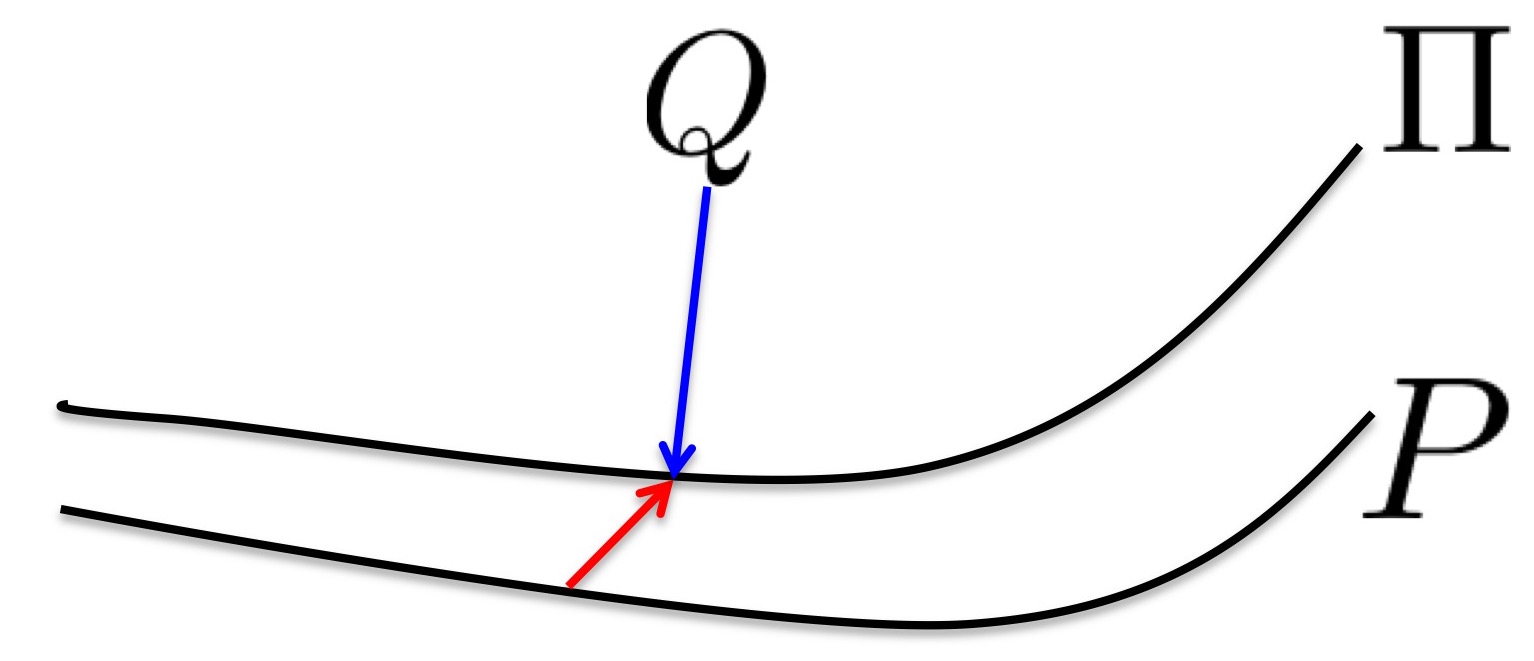} 
	\caption{\small Adversarial contrastive divergence (ACD). Left: Interaction between the models. Red arrow indicates a chasing game, where the generator model chases the energy-based model, which runs toward the data distribution. Right: Contrastive divergence. }	
	\label{fig:t3}
\end{figure}   

Next, consider the learning of the energy-based model model  \cite{Bengio2016, dai2017calibrating}. Recall $S = K - \tilde{K}$ in (\ref{eq:a0}), if we replace the intractable $\pi_{\alpha_t}(x)$ in (\ref{eq:a0}) by $p_\theta(x)$, we get 
\begin{eqnarray} 
\min_\alpha \max_\theta[ {\rm KL}(q_{\rm data}(x)\|\pi_\alpha(x)) - {\rm KL}(p_\theta(x) \| \pi_\alpha(x))],  \label{eq:ACD}
\end{eqnarray} 
or equivalently 
\begin{eqnarray} 
\max_\alpha \min_\theta[ {\rm KL}(p_\theta(x) \| \pi_\alpha(x)) - {\rm KL}(q_{\rm data}(x)\|\pi_\alpha(x))],  \label{eq:ACD1}
\end{eqnarray} 
so that we avoid MCMC for sampling $\pi_{\alpha_t}(x)$, and the gradient for updating $\alpha$ becomes 
\begin{eqnarray}
\frac{\partial}{\partial \alpha} [\E_{q_{\rm data}} (f_\alpha(x)) - \E_{p_\theta}(f_\alpha(x))]. \label{eq:ACD3}
\end{eqnarray}

Because of the negative sign in front of the second KL-divergence in (\ref{eq:ACD}), we need $\max_\theta$ in (\ref{eq:ACD}) or $\min_\theta$ in (\ref{eq:ACD1}), so that the learning becomes adversarial.  See Figure \ref{fig:t3} for illustration. Inspired by {\cite{Hinton2002a}, we call (\ref{eq:ACD}) the adversarial contrastive divergence (ACD). It underlies \cite{Bengio2016, dai2017calibrating}. 

The adversarial form (\ref{eq:ACD}) or (\ref{eq:ACD1}) defines a chasing game with the following dynamics: the generator $p_\theta$ chases the energy-based model $\pi_\alpha$ in $\min_\theta {\rm KL}(p_\theta \| \pi_\alpha)$, the energy-based model $\pi_\alpha$ seeks to get closer to $q_{\rm data}$ and get away from $p_\theta$. The red arrow in Figure \ref{fig:t3} illustrates this chasing game. The result is that $\pi_\alpha$ lures $p_\theta$ toward $q_{\rm data}$. In the idealized case, $p_\theta$ always catches up with $\pi_\alpha$, then $\pi_\alpha$ will converge to the maximum likelihood estimate $\min_\alpha \KL(q_{\rm data}\|\pi_\alpha)$, and $p_\theta$ converges to $\pi_\alpha$.  

The above chasing game is different from VAE $\min_\theta \min_\phi \KL(Q\|P)$, which defines a cooperative game where $q_\phi$ and $p_\theta$ run toward each other. 

Even though the above chasing game is adversarial, both models are running toward the data distribution. While the generator model runs after the energy-based model, the energy-based model runs toward the data distribution. As a consequence, the energy-based model guides or leads the generator model toward the data distribution. It is different from  GAN~\cite{goodfellow2014generative}. In GAN, the discriminator eventually becomes a confused one because the generated data become similar to the real data. In the above chasing game, the energy-based model becomes close to the data distribution. 

The updating of $\alpha$ by (\ref{eq:ACD3}) bears similarity to Wasserstein GAN (WGAN)~\cite{arjovsky2017wasserstein}, but unlike WGAN, $f_\alpha$ defines a probability distribution $\pi_\alpha$, and the learning of $\theta$  is based on $\min_\theta {\rm KL}(p_\theta(x) \| \pi_\alpha(x))$, which is a variational approximation to $\pi_\alpha$. This variational approximation only requires knowing $f_\alpha(x)$, without knowing $Z(\alpha)$. However, unlike $q_\phi(z|x)$, $p_\theta(x)$ is still intractable, in particular, its entropy does not have a closed form. Thus, we can again use variational approximation, by changing the problem to $\min_\theta \min_\phi$$ {\rm KL}(p(z) p_\theta(x|z) \| \pi_\alpha(x) q_\phi(z|x))$, i.e., $\min_\theta \min_\phi {\rm KL}(P\|\Pi)$, which is analytically tractable and which underlies \cite{dai2017calibrating}. In fact, 
\begin{eqnarray}
    {\rm KL}(P\|\Pi) = {\rm KL}(p_\theta(x)\|\pi_\alpha(x)) + {\rm KL}(p_\theta(z|x)\|q_\phi(z|x)). 
\end{eqnarray}
Thus, we can modify (\ref{eq:ACD1}) into $\max_\alpha \min_\theta \min_\phi [{\rm KL}(P\|\Pi) - {\rm KL}(Q \|\Pi)]$, because again ${\rm KL}(Q\|\Pi) = \KL(q_{\rm data}\|\pi_\alpha)$. 

Fitting the above together, we have the divergence triangle (\ref{eq:triangle}), which has a compact symmetric and anti-symmetric form. 

\subsection{Gap Between Two Models}

We can write the objective function ${\cal D}$ as 
\begin{flalign*}
{\cal D}&= ( {\rm KL}(q_{\rm data}(x) \| p_\theta(x)) + {\rm KL}(q_\phi(z|x)\|p_\theta(z|x)))\\
&- ({\rm KL}(q_{\rm data}(x)\|\pi_\alpha(x)) - {\rm KL}(p(z) p_\theta(x|z) \| \pi_\alpha(x) q_\phi(z|x)))\\
&= (( {\rm KL}(q_{\rm data}(x) \| p_\theta(x)) - {\rm KL}(q_{\rm data}(x)\|\pi_\alpha(x)) )\\
&+ {\rm KL}(q_\phi(z|x)\|p_\theta(z|x))+{\rm KL}(p(z) p_\theta(x|z) \| \pi_\alpha(x) q_\phi(z|x)).
\end{flalign*}
Thus ${\cal D}$ is an upper bound of the difference between the log-likelihood of the energy-based model and the log-likelihood of the generator model. 

\subsection{Two Sides of KL-divergences}

In the divergence triangle, the generator model appears on the right side of ${\rm KL}(Q\|P)$, and it also appears on the left side of ${\rm KL}(P\|\Pi)$. The former tends to interpolate or smooth the modes of $Q$, while the later tends to seek after major modes of $\Pi$ while ignoring minor modes. As a result, the learned generator model tends to generate sharper images. As to the inference model $q_\phi(z|x)$, it appears on the left side of ${\rm KL}(Q\|P)$, and it also appears on the right side of ${\rm KL}(P\|\Pi)$. The former is variational learning of the real data, while the latter corresponds to the sleep phase of wake-sleep learning, which learns from the dream data generated by $P$. The inference model thus can infer $z$ from both observed $x$ and generated $x$. 

In fact, if we define 
\begin{eqnarray}
   {\cal D}_0 = {\rm KL}(q_{\rm data}\|p_\theta) + {\rm KL}(p_\theta\|\pi_\alpha) - {\rm KL}(q_{\rm data}\|\pi_\alpha), \label{eq:D0}
\end{eqnarray}
we have 
\begin{eqnarray} 
{\cal D} = {\cal D}_0 + {\rm KL}(q_\phi(z|x) \| p_\theta(z|x))+ {\rm KL}(p_\theta(z|x) \| q_\phi(z|x)). \label{eq:D1}
\end{eqnarray}
(\ref{eq:D0}) is the divergence triangle between the three marginal distributions on $x$, where $p_\theta$ appears on both sides of KL-divergences. (\ref{eq:D1}) is the variational scheme to make the marginal distributions into the joint distributions, which are more tractable. In (\ref{eq:D1}), the two KL-divergences have reverse orders. 

\subsection{Training Algorithm}
The three models are each parameterized by convolutional neural networks. The joint learning under the divergence triangle can be implemented by stochastic gradient descent, where the expectations are replaced by the sample averages. Algorithm~\ref{alg:dt} describes the procedure which is illustrated in Figure~\ref{fig:train}.

\begin{figure}[H]
	\centering	
	\includegraphics[height=.28\linewidth]{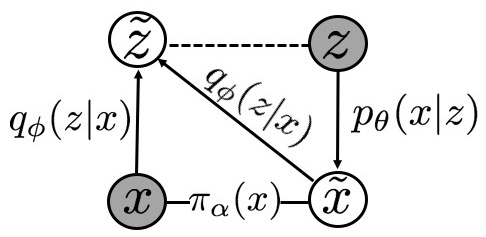} 	
	\caption{Joint learning of three models. The shaded circles $z$ and $x$ represent variables that can be sampled from  the true distributions, i.e., ${\rm N}(0, I_d)$ and empirical data distribution, respectively. $\tilde{x}$ and $\tilde{z}$ are generated samples using the generator model and the inference model, respectively. The solid line with arrow represents the conditional mapping and dashed line indicates the matching loss is involved.}
	\label{fig:train}
\end{figure}

\begin{algorithm}[H]
	\caption{Joint Training for Divergence Triangle Model}
	\label{alg:dt}
	\begin{algorithmic}[1]
		\REQUIRE ~~\\
		training images $\{x_{(i)}\}_{i=1}^{n}$,\\
		number of learning iterations $T$,\\
		$\alpha$, $\theta$, $\phi \leftarrow$ initialized network parameters.  
		\ENSURE~~\\
		estimated parameters $\{\alpha, \theta, \phi\}$,\\
		generated samples $\{\tilde{x}_{(i)}\}_{i=1}^{\tilde{n}}$.	
		\STATE Let $t \leftarrow 0$.
		\REPEAT 
		\STATE $\{z_{(i)} \sim p(z)\}_{i=1}^{\tilde{M}}$.
		\STATE $\{\tilde{x}_{(i)} \sim p_\theta(x|z_{(i)})\}_{i=1}^{\tilde{M}}$.
		\STATE $\{x_{(i)} \sim \q(x)\}_{i=1}^{M}$.
		\STATE $\{\tilde{z}_{(i)} \sim q_\phi(z|x_{(i)})\}_{i=1}^{M}$.
		\STATE {\bf $\alpha$-step}: Given $\{\tilde{x}_{(i)}\}_{i=1}^{\tilde{M}}$ and $\{x_{(i)}\}_{i=1}^{M}$,\\ update $\alpha \leftarrow \alpha + \eta_\alpha \frac{\partial}{\partial \alpha}\D$ with learning rate $\eta_\alpha$. 
		\STATE {\bf $\phi$-step}:  Given $ \{(z_{(i)}, \tilde{x}_{(i)})\}_{i=1}^{\tilde{M}}$ and $ \{(\tilde{z}_{(i)}, x_{(i)})\}_{i=1}^{M}$,\\
		update $\phi \leftarrow \phi - \eta_\phi \frac{\partial}{\partial \phi}\D$, with learning rate $\eta_\phi$. 
		\STATE {\bf $\theta$-step}: Given $ \{(z_{(i)}, \tilde{x}_{(i)})\}_{i=1}^{\tilde{M}}$ and $\{(\tilde{z}_{(i)}, x_{(i)})\}_{i=1}^{M}$,\\
		update $\theta \leftarrow \theta - \eta_\theta \frac{\partial}{\partial \theta}\D $, with learning rate $\eta_\theta$\\
		(optional: multiple-step update).
		\STATE Let $t \leftarrow t+1$.
		\UNTIL $t = T$
	\end{algorithmic}
\end{algorithm}

\section{Experiments}

\begin{table*}
	\begin{center}
		\begin{tabular}{|c|c|c|c|c|c|c|c|c|}
			\hline Model & VAE~\cite{kingma2013auto} & DCGAN~\cite{radford2015unsupervised} & WGAN~\cite{arjovsky2017wasserstein} & CoopNet~\cite{coopnets_pami} & CEGAN~\cite{dai2017calibrating} & ALI~\cite{dumoulin2016adversarially} & ALICE~\cite{li2017alice} & Ours \\
			\hline CIFAR-10 (IS) & 4.08 & 6.16 & 5.76 & 6.55 & 7.07 & 5.93 & 6.02 & \bf{7.23}\\
			\hline CelebA (FID) & 99.09 & 38.39 & 36.36 & 56.57 & 41.89 & 60.29 & 46.14 & \bf{31.92}\\
			\hline
		\end{tabular}
	\end{center}
	\caption{Sample quality evaluation. Row 1: Inception scores for CIFAR-10. Row 2: FID scores for CelebA. }
	\label{tab:is_fid}
\end{table*}

In this section, we demonstrate not only that the divergence triangle is capable of successfully learning an energy-based model with a well-behaved energy landscape, a generator model with highly realistic samples, and an inference model with faithful reconstruction ability, but we also show competitive performance on four tasks: image generation, test image reconstruction, energy landscape mapping, and learning from incomplete images. For image generation, we consider spatial stationary texture images, temporal stationary dynamic textures, and general object categories. We also test our model on large-scale datasets and high-resolution images. 

The images are resized and scaled to $[-1, 1]$, no further pre-processing is needed. The network parameters are initialized with zero-mean Gaussian with standard deviation $0.02$ and optimized using Adam~\cite{kingma2014adam}. Network weights are decayed with rate $0.0005$, and batch normalization~\cite{ioffe2015batch} is used. We refer to the Appendix for the model specifications.

\subsection{Image Generation}

In this experiment, we evaluate the visual quality of generator samples from our divergence triangle model. If the generator model is well-trained, then the obtained samples should be realistic and match the visual features and contents of training images.

\subsubsection{Object Generation}	
  
For object categories, we test our model on two commonly-used datasets of natural images: CIFAR-10 and CelebA~\cite{liu2015faceattributes}. For CelebA face dataset, we randomly select 9,000 images for training and another 1,000 images for testing in reconstruction task. The face images are resized to $64\times 64$ and CIFAR-10 images remain $32 \times 32$. The qualitative results of generated samples for objects are shown in Figure~\ref{fig:generation}. We further evaluate our model using quantitative evaluations which are based on the Inception Score (IS)~\cite{salimans2016improved} for CIFAR-10 and Frechet Inception Distance (FID)~\cite{lucic2017gans} for CelebA faces. We generate 50,000 random samples for the computation of the inception score and 10,000 random samples for the computation of the FID score. Table~\ref{tab:is_fid} shows the IS and FID scores of our model compared with VAE~\cite{kingma2013auto}, DCGAN~\cite{radford2015unsupervised}, WGAN~\cite{arjovsky2017wasserstein}, CoopNet~\cite{coopnets_pami}, CEGAN~\cite{dai2017calibrating}, ALI~\cite{dumoulin2016adversarially}, ALICE~\cite{li2017alice}.

\begin{figure}
	\begin{center}
		\includegraphics[width=0.48\linewidth]{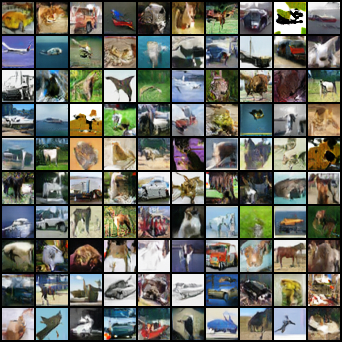}
		\includegraphics[width=0.48\linewidth]{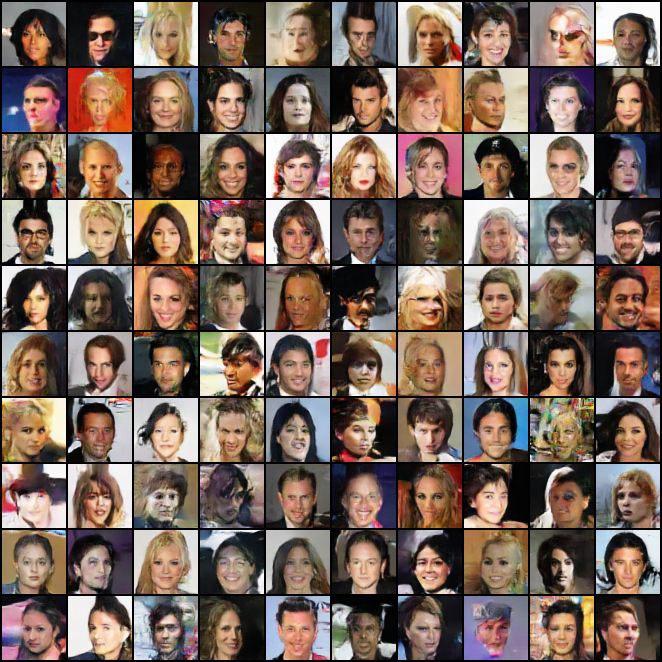}
	\end{center}
	\caption{Generated samples. Left: generated samples on CIFAR-10 dataset. Right: generated samples on CelebA dataset.}
	\label{fig:generation}
\end{figure}

Note that for the Inception Score on CIFAR-10, we borrowed the scores from relevant papers, and for FID score on 9,000 CelebA faces, we re-implemented or used the available code with the similar network structure as our model. It can be seen that our model achieves the competitive performance compared to recent baseline models.

\subsubsection{Large-scale Dataset}

We also train our model on large scale datasets including down-sampled $32\times 32$ version of ImageNet~\cite{oord2016pixel,russakovsky2015imagenet} (roughly 1 million images) and Large-scale Scene Understand (LSUN) dataset~\cite{yu2015lsun}. For the LSUN dataset, we consider the \textit{bedroom}, \textit{tower} and \textit{Church ourdoor} categories which contains roughly 3 million, 0.7 million and 0.1 million images and were re-sized to $64\times 64$. The network structures are similar with the ones used in object generation with twice the number of channels and batch normalization is used in all three models. Generated samples are shown on Figure~\ref{fig:generation_large}.

\begin{figure}
	\begin{center}
	  \includegraphics[width=0.48\linewidth]{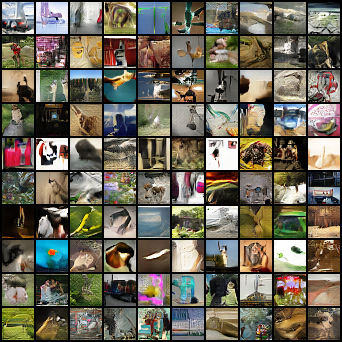}
	   \includegraphics[width=0.48\linewidth]{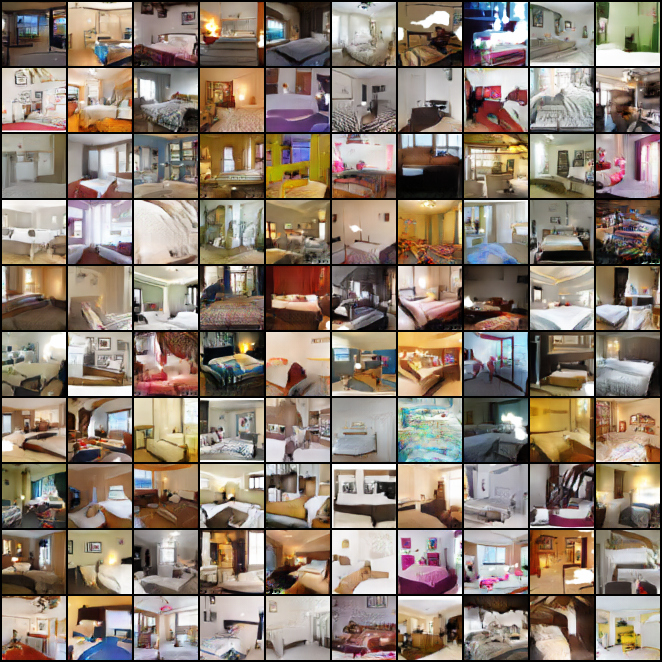}
	  \end{center}
\caption{Generated samples. Left: $32\times 32$ ImageNet. Right: $64 \times 64$ LSUN (bedroom).
}
\label{fig:generation_large}
\end{figure}

\subsubsection{High-resolution Synthesis}

\begin{figure*}
	\begin{center}
		\includegraphics[width=1\linewidth]{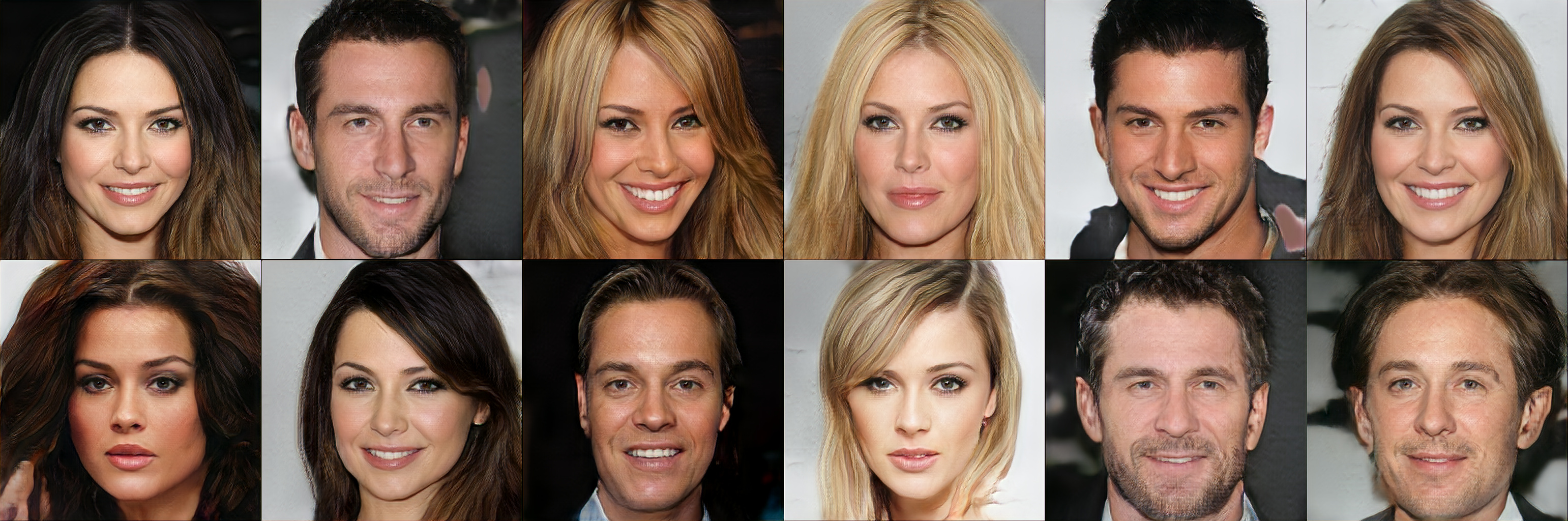}
	\end{center}
	\caption{Generated samples with $1,024\times 1,024$ resolution drawn from $g_\theta(z)$ with 512-dimensional latent vector $z\sim N(0,I_d)$ for CelebA-HQ.}
	\label{fig:celeba_hq}
\end{figure*}

\begin{figure*}
	\begin{center}
		\includegraphics[width=\linewidth]{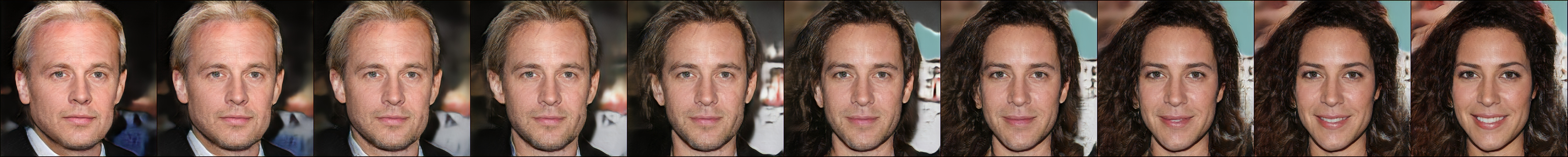}
		\includegraphics[width=\linewidth]{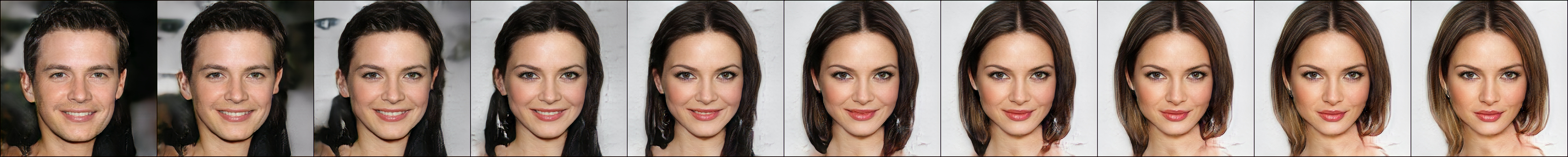}
	\end{center}
	\caption{High-resolution synthesis from the generator model $g_\theta(z)$ with linear interpolation in latent space (i.e., $(1-\alpha)\cdot z_0 + \alpha\cdot z_1$) for CelebA-HQ.}
	\label{fig:celeba_hq_interpolate}
\end{figure*}

In this section, we recruit a layer-wise training scheme to learn models on CelebA-HQ \cite{karras2017progressive} with resolutions of up to $1,024 \times 1,024$ pixels. Layer-wise training dates back to initializing deep neural networks by Restricted Boltzmann Machines to overcome optimization hurdles \cite{hinton2006reducing, bengio2007greedy} and has been resurrected in progressive GANs \cite{karras2017progressive}, albeit the order of layer transitions is reversed such that top layers are trained first. This resembles a Laplacian Pyramid \cite{denton2015deep} in which images are generated in a coarse-to-fine fashion.

As in \cite{karras2017progressive}, the training starts with down-sampled images with a spatial resolution of $4 \times 4$ while progressively increasing the size of the images and number of layers. All three models are grown in synchrony where $1 \times 1$ convolutions project between RGB and feature. In contrast to \cite{karras2017progressive}, we do not require mini-batch discrimination to increase variation of $g_\theta(\cdot)$ nor gradient penalty to preserve $1$-Lipschitz continuity of $f_\alpha(\cdot)$.

Figure~\ref{fig:celeba_hq} depicts high-fidelity synthesis in a resolution of $1,024\times1,024$ pixels sampled from the generator model $g_\theta(z)$ on CelebA-HQ. Figure~\ref{fig:celeba_hq_interpolate} illustrates linear interpolation in latent space (i.e., $(1-\alpha)\cdot z_0 + \alpha\cdot z_1$), which indicates diversity in the samples. 

Therefore, the joint learning in the triangle formulation is not only able to train the three models with stable optimization, but it also achieves synthesis with high fidelity.

\subsubsection{Texture Synthesis}

We consider texture images, which are spatial stationary and contain repetitive patterns. The texture images are resized to $224\times 224$. Separate models are trained on each image. We start from the latent factor of size $7\times7\times5$ and use five convolutional-transpose layers with kernel size $4$ and up-sampling factor $2$ for the generator network. The layers have $512$, $512$, $256$, $128$ and $3$ filters, respectively, and ReLU non-linearity between each layer is used. The inference model has the inverse or ``mirror" structure of generator model except that we use convolutional layers and ReLU with leak factor $0.2$. The energy-based model has three convolutional layers. The first two layers have kernel size $7$ with stride $2$ for $100$ and $70$ filters respectively, and the last layer has $30$ filters with kernel size $5$ and stride $1$.

The representative examples are shown in Figure~\ref{fig:texture}. Three texture synthesis results are obtained by sampling different latent factors from prior distribution $p(z)$. Notice that although we only have one texture image for training, the proposed triangle divergence model can effectively utilize the repetitive patterns, thus generating realistic texture images with different configurations.

\begin{figure}
	\begin{center}
		\includegraphics[width=0.209\linewidth]{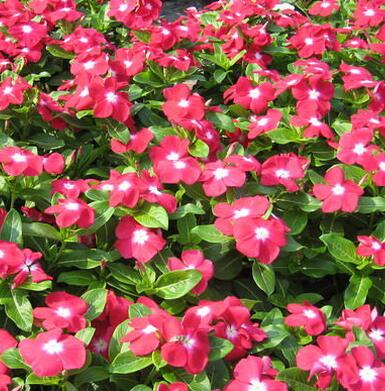}
		\includegraphics[width=0.209\linewidth]{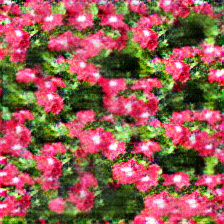}
		\includegraphics[width=0.209\linewidth]{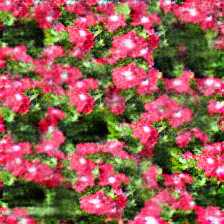}
		\includegraphics[width=0.209\linewidth]{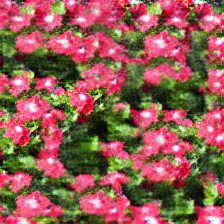}\\
		\vspace{0.5mm}
		\includegraphics[width=0.209\linewidth]{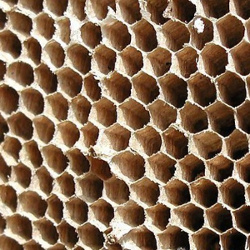}
		\includegraphics[width=0.209\linewidth]{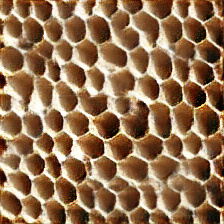}
		\includegraphics[width=0.209\linewidth]{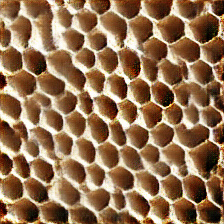}
		\includegraphics[width=0.209\linewidth]{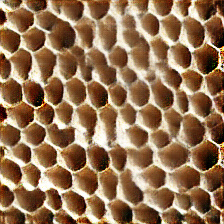}\\
		\vspace{0.5mm}
		\vspace{0.5mm}
		\includegraphics[width=0.209\linewidth]{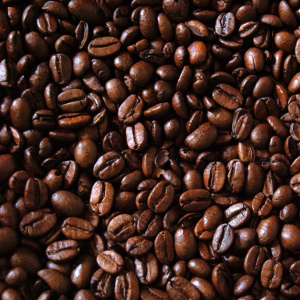}
		\includegraphics[width=0.209\linewidth]{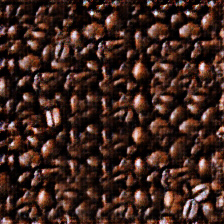}
		\includegraphics[width=0.209\linewidth]{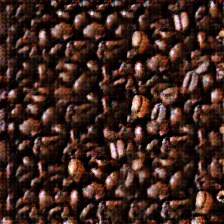}
		\includegraphics[width=0.209\linewidth]{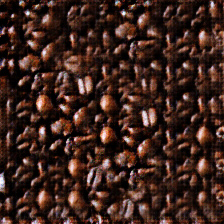}\\
		\vspace{0.5mm}
		\includegraphics[width=0.209\linewidth]{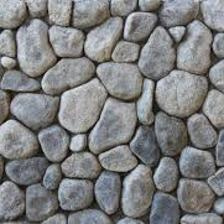}
		\includegraphics[width=0.209\linewidth]{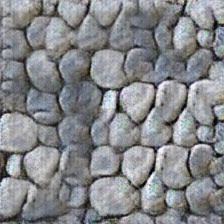}
		\includegraphics[width=0.209\linewidth]{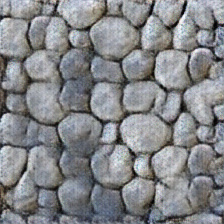}
		\includegraphics[width=0.209\linewidth]{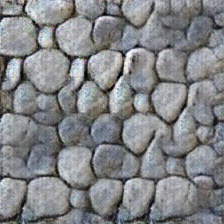}\\
		\vspace{0.5mm}
		\includegraphics[width=0.209\linewidth]{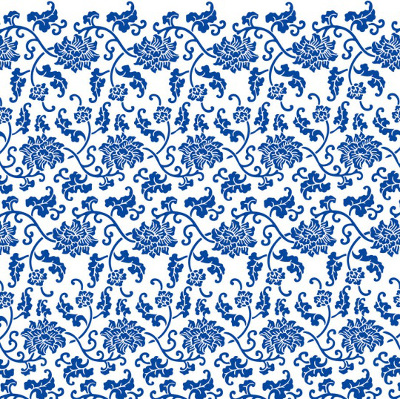}
		\includegraphics[width=0.209\linewidth]{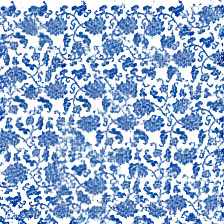}
		\includegraphics[width=0.209\linewidth]{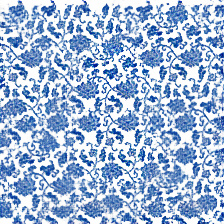}
		\includegraphics[width=0.209\linewidth]{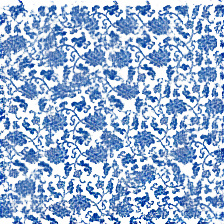}\\
	\end{center}
	\caption{Generated texture patterns. For each row, the left one is the training texture, the remaining images are 3 textures generated by divergence triangle.}
	\label{fig:texture}
\end{figure}

\begin{figure}
	\begin{center}
		\includegraphics[width=0.15\linewidth]{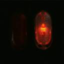}
		\includegraphics[width=0.15\linewidth]{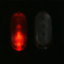}
		\includegraphics[width=0.15\linewidth]{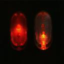}
		\includegraphics[width=0.15\linewidth]{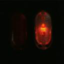}
		\includegraphics[width=0.15\linewidth]{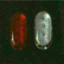}
		\includegraphics[width=0.15\linewidth]{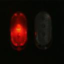}\\
		\vspace{0.5mm}
		\includegraphics[width=0.15\linewidth]{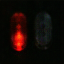}
		\includegraphics[width=0.15\linewidth]{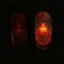}
		\includegraphics[width=0.15\linewidth]{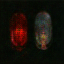}
		\includegraphics[width=0.15\linewidth]{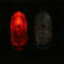}
		\includegraphics[width=0.15\linewidth]{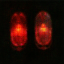}
		\includegraphics[width=0.15\linewidth]{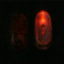}\\
		\vspace{1mm}
		\includegraphics[width=0.15\linewidth]{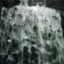}
		\includegraphics[width=0.15\linewidth]{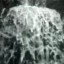}
		\includegraphics[width=0.15\linewidth]{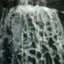}
		\includegraphics[width=0.15\linewidth]{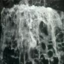}
		\includegraphics[width=0.15\linewidth]{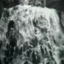}
		\includegraphics[width=0.15\linewidth]{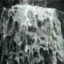}\\
		\vspace{0.5mm}
		\includegraphics[width=0.15\linewidth]{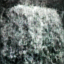}
		\includegraphics[width=0.15\linewidth]{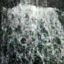}
		\includegraphics[width=0.15\linewidth]{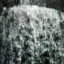}
		\includegraphics[width=0.15\linewidth]{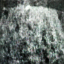}
		\includegraphics[width=0.15\linewidth]{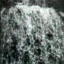}
		\includegraphics[width=0.15\linewidth]{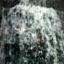}\\
		\vspace{1mm}
		\includegraphics[width=0.15\linewidth]{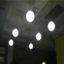}
		\includegraphics[width=0.15\linewidth]{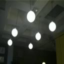}
		\includegraphics[width=0.15\linewidth]{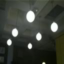}
		\includegraphics[width=0.15\linewidth]{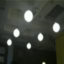}
		\includegraphics[width=0.15\linewidth]{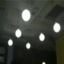}
		\includegraphics[width=0.15\linewidth]{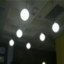}\\
		\vspace{0.5mm}
		\includegraphics[width=0.15\linewidth]{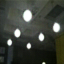}
		\includegraphics[width=0.15\linewidth]{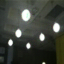}
		\includegraphics[width=0.15\linewidth]{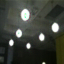}
		\includegraphics[width=0.15\linewidth]{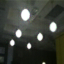}
		\includegraphics[width=0.15\linewidth]{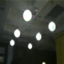}
		\includegraphics[width=0.15\linewidth]{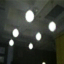}\\
	\end{center}
	\caption{Generated dynamic texture patterns. The top row shows the frames from the training video, the bottom row represents the frames for the generated video.}
	\label{fig:dt}
\end{figure}

\subsubsection{Dynamic Texture Synthesis}

Our model can also be used for dynamic patterns which exhibit stationary regularity in the temporal domain. The training video clips are selected from Dyntex database~\cite{peteri2010dyntex} and resized to $64$ pixels $\times$ $64$ pixels $\times$ $32$ frames. Inspired by recent work~\cite{vondrick2016generating,han2019generator}, we adopt spatial-temporal models for dynamic patterns that are stationary in the temporal domain but non-stationary in the spatial domain. Specifically, we start from $10$ latent factors of size $1\times1\times 2$ for each video clip and we adopt the same spatial-temporal convolutional transpose generator network as in~\cite{han2019generator} except we use kernel size $5$ for the second layer. For the inference model, we use $5$ spatial-temporal convolutional layers. The first $4$ layers have kernel size $4$ with upsampling factor $2$ and the last layer is fully-connected in spatial domain but convolutional in the temporal domain, yielding re-parametrized $\mu_\phi$ and $\sigma_\phi$ which have the same size the as latent factors. For the energy-based model, we use three spatial-temporal convolutional layers. The first two layers have kernel size $4$ with up-sample factor $2$ in all directions, but the last layer is fully-connected in the spatial domain but convolutional with kernel size $4$ and upsample by $2$ in the temporal domain. Each layer has $64$, $128$ and $128$ filters, respectively. Some of the synthesis results are shown in Figure~\ref{fig:dt}. Note, we sub-sampled $6$ frames of the training and generated video clips and we only show them in the first batch for illustration.

\subsection{Test Image Reconstruction}

\begin{table}
	\begin{center}
		\begin{tabular}{|c|c|c|c|c|c|}
			\hline Model & WS~\cite{hinton1995wake} & VAE~\cite{kingma2013auto}  & ALI~\cite{dumoulin2016adversarially} & ALICE~\cite{li2017alice} & Ours \\
			\hline CIFAR-10 & 0.058 & 0.037 & 0.311 &0.034 & \bf{0.028}\\
			\hline CelebA & 0.152 & 0.039 & 0.519 & 0.046  & \bf{0.030}\\
			\hline
		\end{tabular}
	\end{center}
	\caption{Test reconstruction evaluation. Row 1: MSE for CIFAR-10 test set. Row 2: MSE for 1,000 hold out set from CelebA.}
	\label{tab:mse}
\end{table}

\begin{figure}
	\begin{center}
		\includegraphics[width=.48\linewidth]{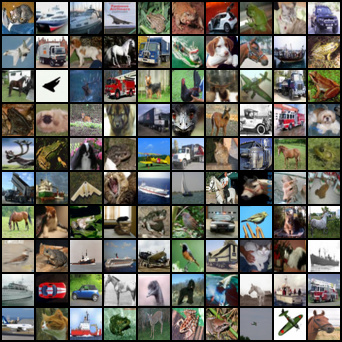}
		\includegraphics[width=0.48\linewidth]{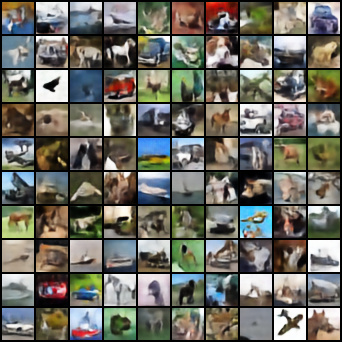}
		\\
		\vspace{1mm}
		\includegraphics[width=0.48\linewidth]{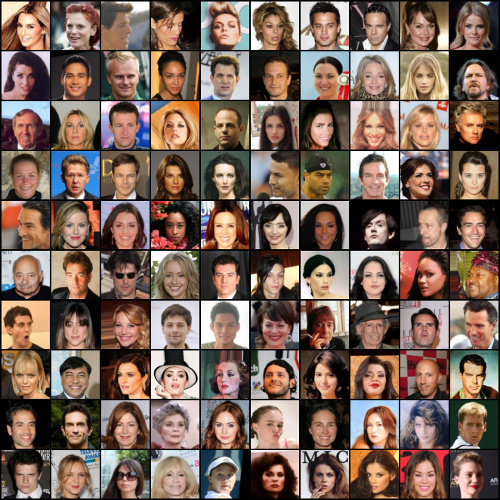}
		\includegraphics[width=0.48\linewidth]{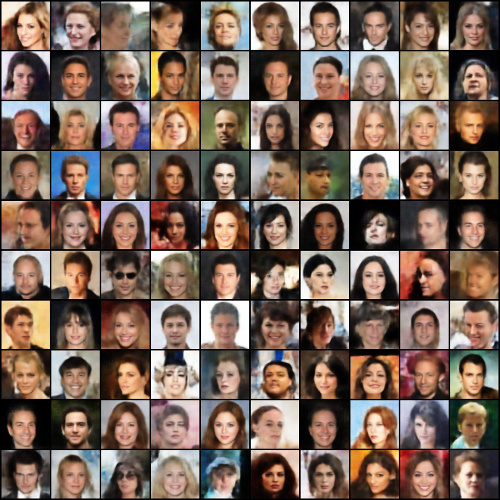}
	\end{center}
	\caption{Test image reconstruction. Top: CIFAR-10. Bottom: CelebA. Left: test images. Right: reconstructed images.}
	\label{fig:test_reconstruction}
\end{figure}

In this experiment, we evaluate the reconstruction ability of our model for a hold-out testing image dataset. This is a strong indicator for the accuracy of our inference model. Specifically, if our divergence triangle model $\D$ is well-learned, then the inference model should match the true posterior of generator model, i.e., $q_\phi(z|x) \approx p_\theta(z|x)$. Therefore, given test signal $x_{te}$, its reconstruction $\tilde{x_{te}}$ should be close to $x_{te}$, i.e., $x_{te}\xrightarrow{q_\phi} z_{te} \xrightarrow{p_\theta} \tilde{x_{te}} \approx x_{te}$. Figure~\ref{fig:test_reconstruction} shows the testing images and their reconstructions on CIFAR-10 and CelebA.

For CIFAR-10, we use its own 10,000 test images while for CelebA, we use the hold-out 1,000 test images as stated above. The reconstruction quality is further measured by per-pixel mean square error (MSE). Table~\ref{tab:mse} shows the per-pixel MSE of our model compared to WS~\cite{hinton1995wake}, VAE~\cite{kingma2013auto}, ALI~\cite{dumoulin2016adversarially}, ALICE~\cite{li2017alice}.

Note, we do not consider methods without inference models on training data, including variants of GANs and cooperative training, since it is infeasible to test such models using image reconstruction.

\subsection{Energy Landscape Mapping}

In the following, we evaluate the learned energy-based model by mapping the macroscopic structure of the energy landscape. When following a MLE regime by minimizing $\KL(\q\|\pi_\alpha)$, we expect the energy-function $-f_\alpha(x)$ to encode ${x\sim \q(x)}$ as local energy minima. Moreover, $-f_\alpha(x)$ should form minima for unseen images and macroscopic landscape structure in which basins of minima are distinctly separated by energy barriers. Hopfield observed that such landscape is a model of associative memory \cite{hopfield1982neural}.

In order to learn a well-formed energy-function, in Algorithm \ref{alg:dt}, we perform multiple $\theta$-steps such that the samples $\{\tilde{x}_i\}_{i=1}^{\tilde{M}}$ are sufficiently ``close'' to the local minima of $-f_\alpha(x)$. This avoids the formation of energy minima not resembling the data. The variational approximation of entropy of the marginal generator distribution $H(p_\theta(x))$ preserves diversity in the samples avoiding mode-collapse.

\begin{figure*}
	\includegraphics[width=1\linewidth, clip, trim={0in 0in 0in 0in}]{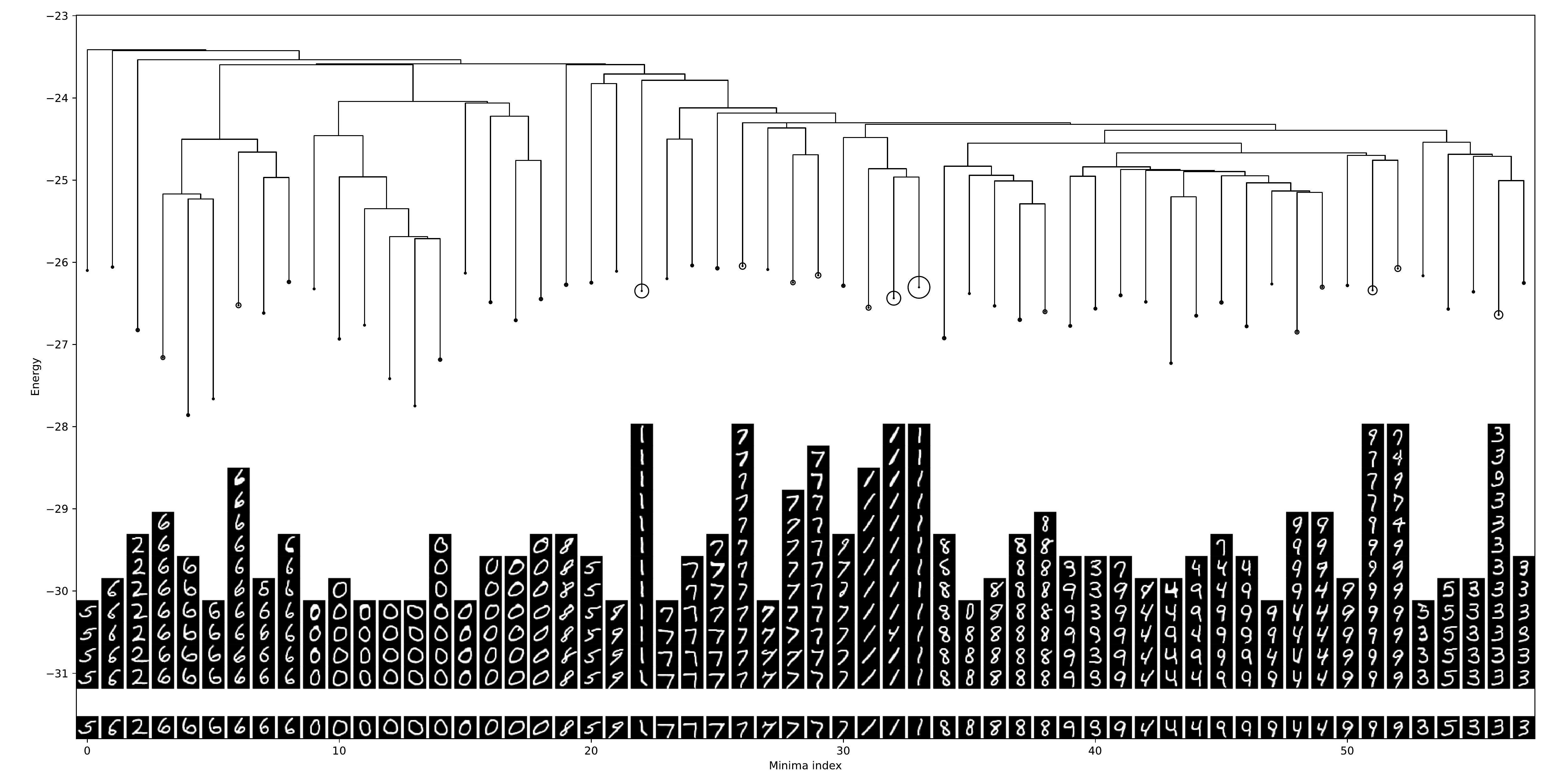}
	\caption{Illustration of the disconnectivity-graph depicting the basin structure of the learned energy-function $f_\alpha(x)$ for the MNIST dataset. Each column represents the set of at most 12 basins members ordered by energy where circles indicate the total number of basin members. Vertical lines encode minima depth in terms of energy and horizontal lines depict the lowest known barrier at which two basins merge in the landscape. Basins with less than 4 members were omitted for clarity.}
	\label{fig:dg_mnist}
\end{figure*}

\begin{figure}
	\includegraphics[width=1.0\linewidth, clip, trim={4.8in 0in 4.8in 0in}]{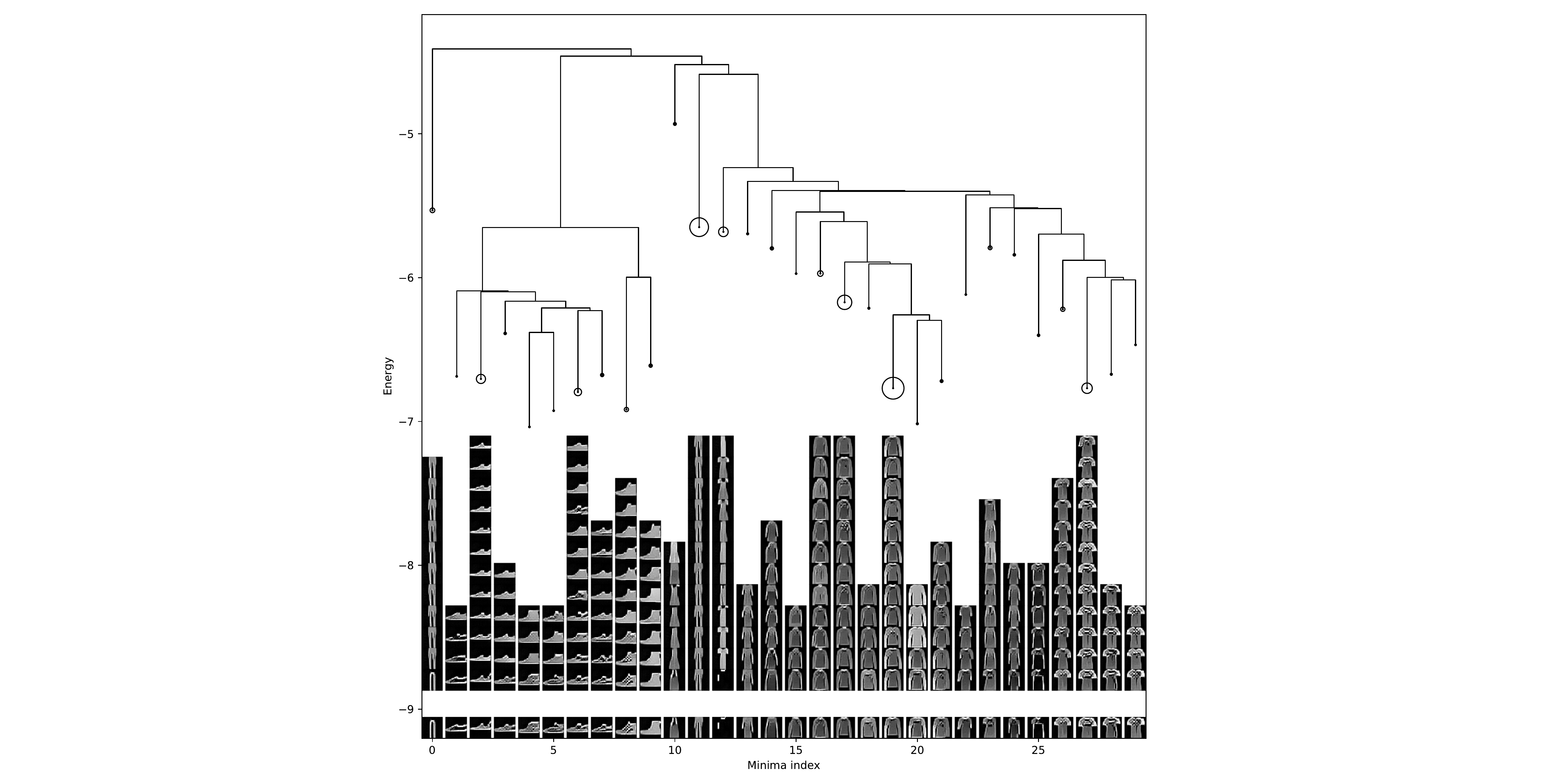}
	\caption{Illustration of the disconnectivity-graph depicting the basin structure of the learned energy-function for the Fashion-MNIST dataset. Each column represents the set of at most 12 basins members ordered by energy where circles indicate the total number of basin members. Vertical lines encode minima depth in terms of energy and horizontal lines depict the lowest known barrier at which two basins merge in the landscape. Basins with less than 4 members were omitted for clarity.}
	\label{fig:dg_fashion}
\end{figure}

\begin{figure}
	\begin{center}
		\includegraphics[width=0.9\linewidth]{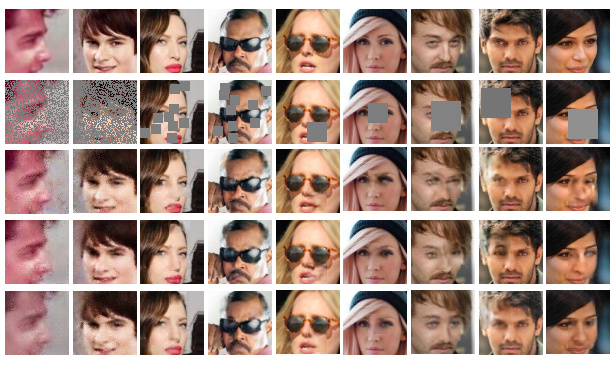}
	\end{center}
	\caption{Learning from incomplete data from the CelebA dataset. The 9 columns belong to experiments P.5, P.7, MB10, MB10, B20, B20, B30, B30, B30 respectively. Row 1: original images, not observed in learning stage. Row 2: training images. Row 3: recovered images using VAE~\cite{kingma2013auto}. Row 4: recovered images using ABP~\cite{HanLu2016}. Row 5: recovered images using our method. }
	\label{fig:learn_incomplete}
\end{figure}

\begin{figure}
	\begin{center}
		\includegraphics[width=0.32\linewidth]{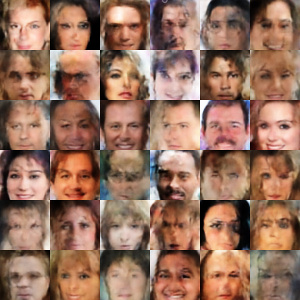}
		\includegraphics[width=0.32\linewidth]{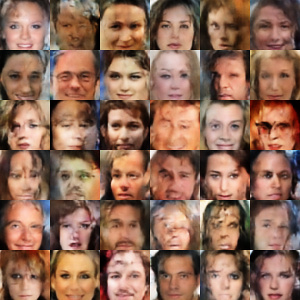}
		\includegraphics[width=0.32\linewidth]{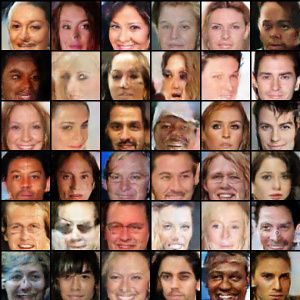}
	\end{center}
	\caption{Image generation from different models learned from training images of the CelebA dataset with $30 \times 30$ occlusions. Left: images generated from VAE model~\cite{kingma2013auto}. Middle: images generated from ABP model~\cite{HanLu2016}. Right: images generated from our proposed triangle divergence model. }
	\label{fig:incomplete_syn}
\end{figure}

To verify that (i) local minima of $-f_\alpha(x)$ resemble $\{x_i\}$ and (ii) minima are separated by significant energy barriers, we shall follow the approach used in~\cite{hill2018building}. When clustering with respect to energetic barriers, the landscape is partitioned into Hopfield basins of attraction whereby each point $\{x_i\}$ on the landscape $-f_\alpha(x)$ is mapped onto a local minimum $\{\hat{x}_i\}$ by a steepest-descent path ${x_i^{t+1} = x_i^{t} + \eta\nabla f_\alpha(x_i^t)}$. The similarity measure used for hierarchical clustering is the barrier energy that separates any two regions. Given a pair of local minima ${\{\hat{x}_i, \hat{x}_j\}}$, we estimate the barrier ${b_{i,j} = \max \{-f_\alpha(x_k) : x_k \in  \hat{x}_i \overset\gamma\rightharpoondown \hat{x}_j\}}$ as the highest energy along a linear interpolation ${x \overset\gamma\rightharpoondown y = \{x + \gamma(y-x) : \gamma \subseteq [0,1]\}}$. If $b_{i,j} < \epsilon$ for some energy threshold $\epsilon$, then ${\{x_i, x_j\}}$ belong to the same basin. The clustering is repeated recursively until all minima are clustered together. Such graphs have come to be referred as disconnectivity graphs (DG)~\cite{wales1998archetypal}.

We conduct energy landscape mapping experiments on the MNIST~\cite{lecun-mnisthandwrittendigit-2010} and Fashion-MNIST~\cite{xiao2017fashion} datasets, each containing $70,000$ grayscale images of size $28\times28$ pixels depicting handwritten digits and fashion products from $10$ categories, respectively. 
The energy landscape mapping is not without limitations, because it is practically impossible to locate all local modes. Based on the local modes located by our algorithm, see Figure~\ref{fig:dg_mnist} for the MNIST dataset, it suggests that the learned energy function is well-formed which not only encodes meaningful images as minima, but also forms meaningful macroscopic structure. Moreover, within basins the local minima have a high degree of purity (i.e. digits within a basin belong to the same class), and, the energy barrier between basins seem informative (i.e. basins of ones and sixes form pure super-basins). Figure~\ref{fig:dg_fashion} depicts the energy landscape mapping on Fashion-MNIST.

Potential applications include unsupervised classification in which energy barriers act as a geodesic similarity measure which captures perceptual distance (as opposed to e.g. $\ell_2$ distance), weakly-supervised classification with one label per basins, or, reconstruction of incomplete data (i.e. Hopfield content-addressable memory or image inpainting).

\subsection{Learning from incomplete images}

The divergence triangle can be used to learn from occluded images. This task is challenging~\cite{HanLu2016}, because only parts of the images are observed, thus the model needs to learn sufficient information to recover the occluded parts.
The generative models with inferential mechanism can be used for this task. Notably,~\cite{HanLu2016} proposed to recover incomplete images using alternating back-propagation (ABP) which has a MCMC based inference step to refine the latent factors and perform reconstruction iteratively. VAEs~\cite{rezende2014stochastic,kingma2013auto} build the inference model on occluded images, and can also be adapted for this task. It proceeds by filling the missing parts with average pixel intensity in the beginning, then iteratively re-update the missing parts using reconstructed values. Unlike VAEs, which only consider the un-occluded parts of training data, our model utilizes the generated samples which become gradually recovered during training, resulting in improved recovery accuracy and sharp generation. Note that learning from incomplete data can be difficult for variants of GANs~\cite{goodfellow2014generative,dai2017calibrating,radford2015unsupervised,arjovsky2017wasserstein} and cooperative training~\cite{coopnets_pami}, since inference cannot be performed directly on the occluded images. 

We evaluate our model on 10,000 images randomly chosen from CelebA dataset. Then, selected images are further center cropped as in~\cite{HanLu2016}. Similar to VAEs, we zero-fill the occluded parts in the beginning, then iterative update missing values using reconstructed images obtained from the generator model. Three types of occlusions are used: (1) salt and pepper noise which randomly covers $50\%$ (P.5) and $70\%$ (P.7) of the image. (2) Multiple block occlusion which has 10 random blocks of size $10 \times 10$ (MB10). (3) Singe block occlusion where we randomly place a large $20\times 20$ and $30 \times 30$ block on each image, denoted by B20 and B30 respectively. Table~\ref{tab:recover} shows the recovery errors using VAE~\cite{kingma2013auto}, ABP~\cite{HanLu2016} and our triangle model where the error is defined as per-pixel absolute difference (relative to the range of pixel values) between the recovered image on the occluded pixels and the ground truth image.

\begin{table}[H]
	\begin{center}
		\begin{tabular}{|c|c|c|c|c|c|}
			\hline EXP & P.5 & P.7 & MB10 & B20 & B30 \\
			\hline VAE~\cite{kingma2013auto} & 0.0446 & 0.0498 &0.1169 & 0.0666 & 0.0800\\
			\hline ABP~\cite{HanLu2016} & \bf{0.0379} & \bf{0.0428} & 0.1070& 0.0633 & 0.0757 \\
			\hline Ours &0.0380 & 0.0430 &\bf{0.1060}& \bf{0.0621} & \bf{0.0733}\\
			\hline
		\end{tabular}
	\end{center}
	\caption{Recovery errors for different occlusion masks for $10,000$ images selected from the CelebA dataset. }
	\label{tab:recover}
\end{table}

It can be seen that our model consistently out-performs the VAE model for different occlusion patterns. For structured occlusions (i.e., multiple and single blocks), the un-occluded parts contain more meaningful configurations that will improve learning of the generator through the energy-based model, which will, in turn, generate more meaningful samples to refine our inference model. This could be verified by the superior results compared to ABP~\cite{HanLu2016}. While for unstructured occlusions (i.e., salt and pepper noise), ABP achieves improved recovery, a possible reason being that un-occluded parts contain less meaningful patterns which offer limited help for learning the generator and inference model. Our model synthesizes sharper and more realistic images from the generator on occluded images. See Figure~\ref{fig:incomplete_syn} in which images are occluded with $30 \times 30$ random blocks. 

\section{Conclusion}

The proposed probabilistic framework, namely divergence triangle, for joint learning of the energy-based model, the generator model, and the inference model. The divergence triangle forms the compact learning functional for three models and naturally unifies aspects of maximum likelihood estimation~\cite{HanLu2016, coopnets_pami}, variational auto-encoder~\cite{kingma2013auto, RezendeICML2014, MnihGregor2014}, adversarial learning~\cite{Bengio2016, dai2017calibrating}, contrastive divergence~\cite{hinton}, and the wake-sleep algorithm~\cite{hinton1995wake}. 

An extensive set of experiments demonstrated learning of a well-behaved energy-based model, realistic generator model as well as an accurate inference model. Moreover, experiments showed that the proposed divergence framework can be effective in learning directly from incomplete data.

In future work, we aim to extend the formulation to learn interpretable generator and energy-based models with multiple layers of sparse or semantically meaningful latent variables or features \cite{salakhutdinov2009deep, lee2009convolutional}. Further, it would be desirable to unify the generator, energy-based and inference models into a single model \cite{dinh2014nice, dinh2016density} by allowing them to share parameters and nodes instead of having separate sets of parameters and nodes.

\ifCLASSOPTIONcompsoc
\section*{Acknowledgments}
\else
\section*{Acknowledgment}
\fi

The work is supported by DARPA XAI project N66001-17-2-4029; ARO project W911NF1810296; and ONR MURI project N00014-16-1-2007; and Extreme Science and Engineering Discovery Environment (XSEDE) grant ASC170063. We thank Dr. Tianfu Wu, Shuai Zhu and Bo Pang for helpful discussions.

\appendix  

%

\section*{Model Architecture}

We describe the basic network structures, in particular for object generation. We use the following notation:

\begin{itemize}
	\item conv(n): convolutional operation with $n$ output feature maps.
	\item convT(n): convolutional transpose operation with $n$ output feature maps. 
	\item LReLU: Leaky-ReLU nonlinearity with default leaky factor 0.2.
	\item BN: Batch normalization.
\end{itemize}

The structures for CelebA (where 9,000 random images are chosen) are shown in Table~\ref{tab:celeba}. The structures for CIFAR-10 and MNIST/Fashion-MNIST are shown in Table~\ref{tab:cifar10} and Table~\ref{tab:mnist}, respectively. 

\begin{table}[H]
	\begin{center}
		\begin{tabular}{ |c|c|c|c| }
			\hline
			\multicolumn{4}{ |c| }{Generator Model} \\
			\hline
			Layers & In-Out Size & Stride  & BN\\ \hline
			Input: Z & 1x1x100 & & \\
			4x4 convT(512), ReLU & 4x4x512 & 1 & Yes \\
			4x4 convT(512), ReLU & 8x8x512 & 2& Yes\\
			4x4 convT(256), ReLU & 16x16x256 & 2& Yes \\
			4x4 convT(128), ReLU & 32x32x128 & 2 & Yes \\ 
			4x4 convT(3), ReLU & 64x64x3 & 2 & No\\
			\hline
			\multicolumn{4}{ |c| }{Inference model} \\
			\hline
			Input: X & 64x64x3 & & \\
			4x4 conv(64), LReLU & 32x32x64 & 2 & Yes \\
			4x4 conv(128), LReLU & 16x16x128 & 2& Yes\\
			4x4 conv(256), LReLU & 8x8x256 & 2& Yes \\
			4x4 conv(512), LReLU & 4x4x512 & 2 & Yes \\ 
			4x4 conv(100), LReLU & $\mu , \sigma$: 1x1x100 & 1 & No\\
			\hline
			\multicolumn{4}{ |c| }{Energy-based Model} \\
			\hline
			Input: X & 64x64x3 & & \\
			4x4 conv(64), LReLU & 32x32x64 & 2 & Yes \\
			4x4 conv(128), LReLU & 16x16x128 & 2& Yes\\
			4x4 conv(256), LReLU & 8x8x256 & 2& Yes \\
			4x4 conv(256), LReLU & 4x4x256 & 2 & Yes \\ 
			4x4 conv(1), LReLU & 1x1x1 & 1 & No\\
			\hline
		\end{tabular}
	\end{center}
	\caption{Network structures for CelebA (9,000).}
	\label{tab:celeba}
\end{table}

\vspace{-1em}

\begin{table}[H]
	\begin{center}
		\begin{tabular}{ |c|c|c|c| }
			\hline
			\multicolumn{4}{ |c| }{Generator Model} \\
			\hline
			Layers & In-Out Size & Stride  & BN\\ \hline
			Input: Z & 1x1x100 & & \\
			4x4 convT(512), ReLU & 4x4x512 & 1 & Yes \\
			4x4 convT(512), ReLU & 8x8x512 & 2& Yes\\
			4x4 convT(256), ReLU & 16x16x256 & 2& Yes \\
			4x4 convT(128), ReLU & 32x32x128 & 2 & Yes \\ 
			3x3 convT(3), Tanh & 32x32x3 & 1 & No\\
			\hline
			\multicolumn{4}{ |c| }{Inference model} \\
			\hline
			Input: X & 32x32x3 & & \\
			3x3 conv(64), LReLU & 32x32x64 & 1 & No \\
			4x4 conv(128), LReLU & 16x16x128 & 2& No\\
			4x4 conv(256), LReLU & 8x8x256 & 2& No \\
			4x4 conv(512), LReLU & 4x4x512 & 2 & No \\ 
			4x4 conv(100) & $\mu , \sigma$: 1x1x100 & 1 & No\\
			\hline
			\multicolumn{4}{ |c| }{Energy-based Model} \\
			\hline
			Input: X & 32x32x3 & & \\
			3x3 conv(64), LReLU & 32x32x64 & 1 & No \\
			4x4 conv(128), LReLU & 16x16x128 & 2& No\\
			4x4 conv(256), LReLU & 8x8x256 & 2& No \\
			4x4 conv(256), LReLU & 4x4x256 & 2 & No \\ 
			4x4 conv(1) & 1x1x1 & 1 & No\\
			\hline
		\end{tabular}
	\end{center}
	\caption{Network structures for CIFAR-10.}
	\label{tab:cifar10}
\end{table}

\vspace{-1em}

\begin{table}[H]
	\begin{center}
		\begin{tabular}{ |c|c|c|c| }
			\hline
			\multicolumn{4}{ |c| }{Generator Model} \\
			\hline
			Layers & In-Out Size & Stride  & BN\\ \hline
			Input: Z & 1x1x100 & & \\
			3x3 convT(1024), ReLU & 3x3x1024 & 1 & Yes \\
			4x4 convT(512), ReLU & 7x7x512 & 2 & Yes\\
			4x4 convT(256), ReLU & 14x14x256 & 2& Yes \\
			4x4 convT(1), Tanh & 28x28x1 & 2 & No\\
			\hline
			\multicolumn{4}{ |c| }{Inference model} \\
			\hline
			Input: X & 28x28x1 & & \\
			4x4 conv(128), LReLU & 14x14x128 & 2 & No \\
			4x4 conv(256), LReLU & 7x7x256 & 2& No\\
			4x4 conv(512), LReLU & 3x3x512 & 2& No \\
			4x4 conv(100) & $\mu , \sigma$: 1x1x100 & 1 & No\\
			\hline
			\multicolumn{4}{ |c| }{Energy-based Model} \\
			\hline
			Input: X & 28x28x1 & & \\
			4x4 conv(128), LReLU & 14x14x128 & 2 & No \\
			4x4 conv(256), LReLU & 7x7x256 & 2& No\\
			4x4 conv(512), LReLU & 3x3x512 & 2& No \\
			4x4 conv(1) & 1x1x1 & 1 & No\\
			\hline
		\end{tabular}
	\end{center}
	\caption{Network structures for MNIST and Fashion-MNIST.}
	\label{tab:mnist}
\end{table}

\ifCLASSOPTIONcaptionsoff
  \newpage
\fi



%

\bibliographystyle{IEEEtran}
\bibliography{triangle}



%








\end{document}